\documentclass[letterpaper, 10 pt, journal, twoside]{ieeetran}
\usepackage{amsmath,amsfonts}
\usepackage{algorithmic}
\usepackage{algorithm}
\usepackage{array}
\usepackage{textcomp}
\usepackage{stfloats}
\usepackage{url}
\usepackage{verbatim}
\usepackage{graphicx}
\usepackage{cite}
\usepackage{picinpar}
\usepackage{url}
\usepackage{flushend}
\usepackage[latin1]{inputenc}
\usepackage{colortbl}
\usepackage{soul}
\usepackage{multirow}
\usepackage{pifont}
\usepackage{color}
\usepackage[table,xcdraw]{xcolor}
\usepackage{alltt}
\usepackage[hidelinks]{hyperref}
\usepackage{enumerate}
\usepackage{siunitx}
\usepackage{breakurl}
\usepackage{epstopdf}
\usepackage{pbox}
\usepackage{gensymb}
\usepackage{amsfonts}
\usepackage{makecell}
\usepackage{subfigure}
\usepackage{mathrsfs}
\usepackage{array}
\usepackage{threeparttable}
\usepackage{amssymb}
\definecolor{mycyan}{cmyk}{.15,0,0,0}

\author{Yanmei Jiao$^{1}$, Binxin Zhang$^{1}$, Peng Jiang$^{1}$, Chaoqun Wang$^{2}$, Rong Xiong$^{3}$ and Yue Wang$^{3}$%
\thanks{$^{1}$Yanmei Jiao, Binxin Zhang, Peng Jiang are with the School of Information Science and Engineering, Hangzhou Normal University, Hangzhou, P.R. China. }%
\thanks{$^{2}$Chaoqun Wang is with the School of Control Science and Engineering, Shandong University, Shandong, P.R. China. }
\thanks{$^{3}$Yue Wang and Rong Xiong are with the State Key Laboratory of Industrial Control and Technology, Zhejiang University, Hangzhou, P.R. China. (Corresponding author: Yue Wang, Email: {\tt\small ywang24@zju.edu.cn}).}%
}

\title{\LARGE \bf
3D Model-free Visual Localization System from Essential Matrix \\ under Local Planar Motion
}

\begin{document}

\maketitle

\begin{abstract}
Visual localization plays a critical role in the functionality of low-cost autonomous mobile robots. Current state-of-the-art approaches for achieving accurate visual localization are 3D scene-specific, requiring additional computational and storage resources to construct a 3D scene model when facing a new environment. An alternative approach of directly using a database of 2D images for visual localization offers more flexibility. However, such methods currently suffer from limited localization accuracy. In this paper, we propose an accurate and robust multiple checking-based 3D model-free visual localization system to address the aforementioned issues. To ensure high accuracy, our focus is on estimating the pose of a query image relative to the retrieved database images using 2D-2D feature matches. Theoretically, by incorporating the local planar motion constraint into both the estimation of the essential matrix and the triangulation stages, we reduce the minimum required feature matches for absolute pose estimation, thereby enhancing the robustness of outlier rejection. Additionally, we introduce a multiple-checking mechanism to ensure the correctness of the solution throughout the solving process. For validation, qualitative and quantitative experiments are performed on both simulation and two real-world datasets and the experimental results demonstrate a significant enhancement in both accuracy and robustness afforded by the proposed 3D model-free visual localization system.
\end{abstract}

\def\abstractname{Note to Practitioners}
\begin{abstract}
The motivation of this article stems from the need to develop an accurate visual localization system with simplicity and flexibility of map construction and easy adaption to new environments.  Such a system holds great practical value for a range of applications, including warehouse robots, service robots, and countless others. Existing visual localization systems that achieve high accuracy are dependent on a pre-built accurate 3D scene map, which pose challenges in terms of map construction and consume significant storage resources onboard, particularly for large scenes. And the aforementioned efforts need to be repeated when changing to a new scene. In this article, an accurate and robust 3D model-free visual localization system is proposed to handle this problem. The map construction is simplified to build a set of database images with associated camera poses, which is trivial as it amounts to adding posed images to a database. The core idea for achieving high accuracy and robustness is to model the local planar motion characteristic of general ground-moving robots into both essential matrix estimation and triangulation stages to obtain two minimal solutions. The proposed localization system simplifies the task of switching between different application scenarios for the robot, reducing additional workload and lowering the difficulty of use.
\end{abstract}


\section{Introduction}
\IEEEPARstart{G}{iven} a query image, visual localization refers to the problem of estimating the pose, \textit{i.e.}, position and orientation of the camera using visual information with respect to a environment. Visual localization is a core functionality for low-cost positioning in autonomous unmanned systems, which has important practical applications in various scenarios such as navigation, exploration, surveillance, and rescue \cite{cheng2019robust, cheng2020improving, li2017indoor, santoso2016visual, jiao2021deterministic,zhang2023toward}.

The most accurate visual localization methods currently available are based on 3D information. The 3D scene information can be either explicit point cloud maps constructed by SfM (Structure-from-Motion) methods\cite{schonberger2016structure}, or implicit scene encoded by neural networks\cite{kendall2015posenet, brachmann2017dsac, balntas2018relocnet}. The general pipeline of 3D scene based visual localization approaches is to first obtain the 2D-3D feature matching between a query image and a pre-built 3D scene\cite{lowe2004distinctive, detone2018superpoint, rublee2011orb}, and then perform accurate localization based on absolute pose estimation algorithms\cite{jiao20202, gao2003complete, lepetit2009epnp}. These methods solve for high pose accuracy, but they are scene-specific, \textit{i.e.}, they require extra time and computational resources to build a high precision 3D map when facing a new scene.

\begin{figure}[tbp]
    \centering
    \includegraphics[width=0.5\textwidth]{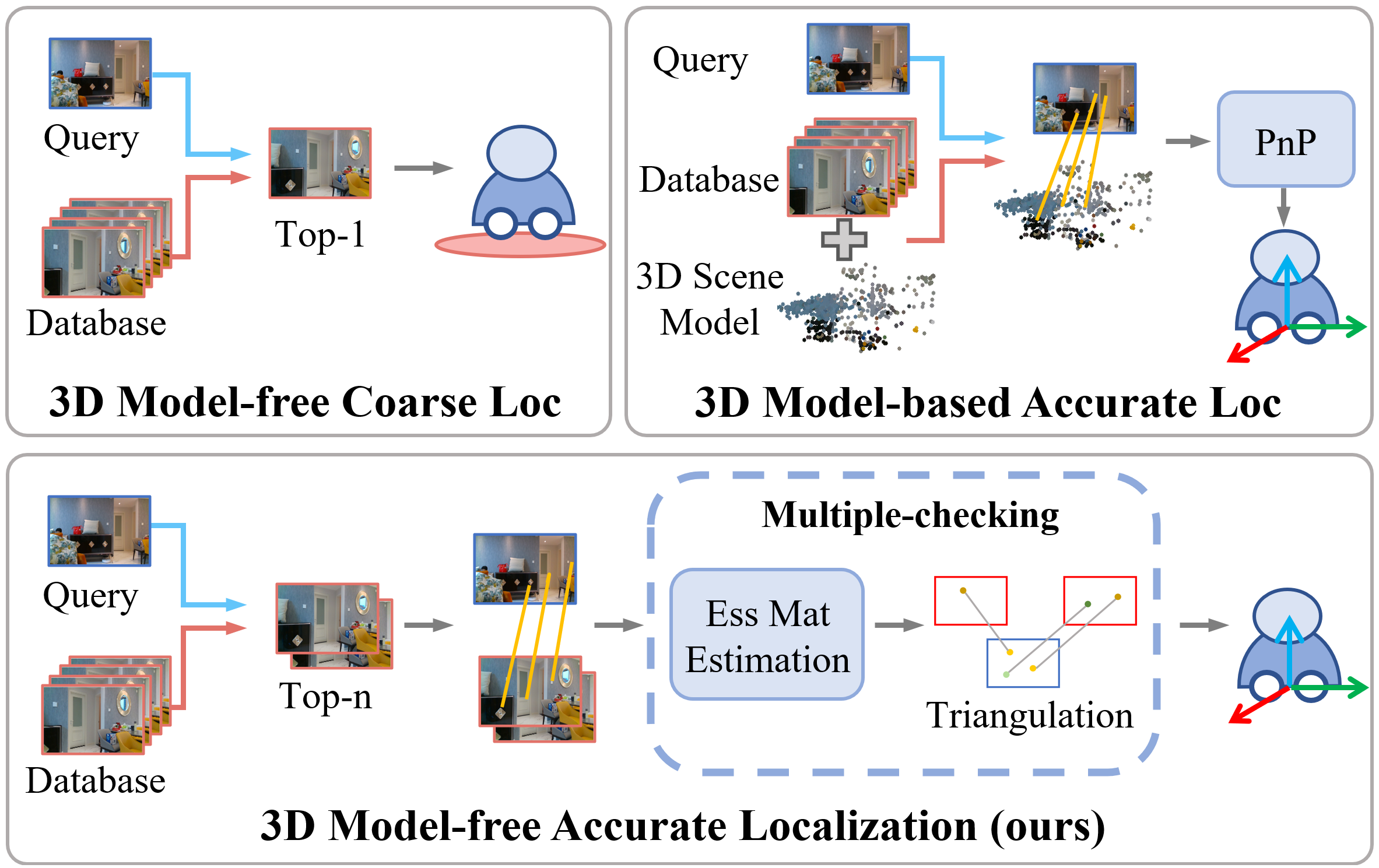}
    \caption{Comparison of different problem settings of visual localization with or without a 3D scene model. The proposed 3D model-free visual localization system achieves accurate and robust localization directly based on a 2D image database, which (1) differentiates itself from a 3D model-based pipeline by eliminating the need for prior construction of a 3D environmental model, (2) achieves higher accuracy compared to typical 3D model-free method using the pose of retrieved database image for approximation, (3) leverages a multiple-checking procedure to ensure the correctness and robustness of essential matrix estimation and triangulation for absolute pose estimation.    }
    \label{fig.overview}
\end{figure}

A more flexible way of visual localization is to localize directly based on a 2D scene database which is a set of reference images with associated camera poses. Currently, this line of method focuses on image retrieval or place recognition, which retrieves the most similar database image and approximates the pose of query image using the associated pose in database \cite{arandjelovic2016netvlad, lowry2015visual, masone2021survey}. However, the localization accuracy of this method is limited by the sampling interval when constructing the database, which is difficult to meet the requirements of tasks that require higher localization accuracy. To further improve the localization accuracy, the pose transformation between the query image and the database image can be estimated from 2D-2D feature matches, which is the focus of this paper.

There are two problems with the existing 2D-2D pose estimation methods for visual localization. First, these methods mostly perform relative pose estimation between two images based on essential matrix estimation, which provides the rotation matrix and the relative translation direction but not the absolute translation scale\cite{hartley2003multiple, nister2004efficient, stewenius2006recent, fischler2014readings}. Therefore, they cannot obtain the absolute pose of the camera with respect to the global coordinate system of the database, which is essential for complete visual localization. Second, most of them formulate the relative pose estimation as a 6DoF (degree-of-freedom) problem, which ignores the prior kinematic constraints in robotics. Mobile robots and ground vehicles usually operate on planar or at least locally planar surfaces such as floors (indoor) and roads (outdoor). The planar motion constraint can simplify the 6DoF relative pose problem to a 3DoF one, which leads to less computation and higher accuracy.

In this paper, we propose two minimal solutions for absolute pose estimation from essential matrix and a multiple checking scheme for robust 3D model-free visual localization, addressing the above problems. The proposed minimal solutions are derived from simplified epipolar geometry that leverages local planar motion constraint. Specifically, based on the simplified epipolar constraint, only two feature point matches are required for essential matrix estimation. Then, a 2p2p minimal solution is naturally proposed, which estimates the absolute pose by triangulating the translation vector obtained from two sets of point correspondences provided by two reference images in the database. However, the 2p2p minimal solution still has redundancy in the constraints, and the number of feature matches required can be further reduced by considering the motion property into triangulation. Therefore, we propose a 2p1p minimal solution that only needs one additional feature match from the second reference image to uniquely determine the absolute translation scale and complete the absolute pose estimation. The proposed minimal solutions are embedded with a multiple checking scheme to perform a robust pose estimation. Note that reducing the number of necessary correspondences for pose estimation is meaningful, as it leads to higher success rate with fixed number of iterations in RANSAC, which means higher robustness of visual localization. In addition, by incorporating affine-invariant feature matching from our previous work \cite{jiao2021deterministic}, which has been reformulated with local planar motion property, a complete and robust 3D model-free visual localization system is present. Extensive simulation and real world experiments are conducted to demonstrate its effectiveness. In summary, the contributions of this paper are listed as follows:
\begin{itemize}
\item We incorporate the local planar motion constraint into both the essential matrix estimation and triangulation stages, resulting in two minimal solutions namely 2p2p and 2p1p for absolute pose estimation using 2D-2D matches.
\item We integrate the proposed minimal solutions with a multiple checking-based mechanism to perform outlier rejection in the early stage and ensure the correctness of pose estimation.
\item We propose a comprehensive 3D model-free visual localization system that combines the affine-invariant feature matching and multiple checking-based pose estimation, operating under local planar motion.
\item We conduct extensive experiments on synthetic data and two real world datasets to validate the accuracy and robustness of the proposed system for visual localization of ground moving robots.
\end{itemize}

\section{Related Work}
\subsection{3D Model-based Visual Localization}
Generally, there are two lines of 3D model-based visual localization approaches. The first is direct approaches, which directly generate 2D-3D feature correspondences from a query image and a 3D scene model \cite{cao2014minimal, choudhary2012visibility, geppert2019efficient}. The typical methods match feature descriptors extracted from a 2D query image and 3D points in a SfM model and then produce accurate camera pose estimation from the matches. While achieving high accuracy, these methods struggle in scaling to large scenes due to memory consumption or arising ambiguities \cite{li2016worldwide}. Recently, inspired by the success of deep learning in image understanding, several works adopt deep neural networks to directly learn 2D-3D feature matching function \cite{brachmann2017dsac, brachmann2018learning, donoser2014discriminative} or regress camera pose by using 2D-3D matches as part of the loss function \cite{kendall2017geometric}. These methods utilize CNNs (convolutional neural network) to autonomously discover useful association information and can be performed end-to-end. However, their scalability to larger scenes is also limited and their generalization ability to new scenes has to be improved. The second is indirect approaches\cite{sarlin2019coarse, irschara2009structure, sarlin2018leveraging}, which follow a hierarchical localization pipeline by using image retrieval methods to retrieve a similar database image and match 2D-2D feature descriptors between query image and database image. The 2D-3D matches are then obtained by retrieving the associated 3D points information of the database image stored in the scene model. Similar to direct methods, these methods are scene-specific which require extra computational resources and efforts to build a high precision 3D map when facing a new scene. While the proposed method in this paper aims to achieve accurate and robust visual localization without a pre-built 3D scene model.

\subsection{3D Model-free Visual Localization}
Visual localization without 3D scene models refers to recovering the pose of a query image from a database of 2D images associated with pose information. It generally includes three steps: image retrieval, relative pose estimation, and absolute pose estimation. Image retrieval, also known as place recognition \cite{arandjelovic2016netvlad, lowry2015visual, masone2021survey}, is the same as the first step of the indirect 3D model-based methods, which is based on image-level descriptors to extract top-$n$ similar reference images from the database.

For each retrieved reference image, the second step is to compute the essential matrix via relative pose estimation algorithms. Relative pose estimation between two images is one of the classical geometric problems in computer vision. For the general 6DoF estimation problem, 8-point algorithm is the most common and standard method \cite{hartley2003multiple, fischler2014readings}. And the minimal solver is 5-point algorithm which performs pose estimation using the least number of correspondences \cite{nister2004efficient, stewenius2006recent}. By considering motion property in robotics, many efforts have been made to further reduce the number of features to solve the simplified relative pose estimation problems. As mobile robots and vehicles usually move on floors or roads, the problem can be reduced to 3DoF. Then 3-point algorithm formulated by a linear equation and the minimal solver based on iterative 2-point algorithm are proposed \cite{ortin2001indoor}. Then the non-iterative 2-point algorithm formulated by finding intersections of two ellipses is proposed \cite{chou20152} and be improved with less computation in \cite{hong2016improved}. With the assumption of planar and circular motion, the 1-point minimal solver is proposed using only a single correspondence \cite{scaramuzza2009real}. However, the above algorithms only estimates the essential matrix between two images, which can be used to recover the rotation matrix and translation direction, but not the translation scale. The proposed minimal solutions in this paper aim to solve the absolute pose between the query image and the 2D image database, which can be directly used for visual localization.

The final step of absolute pose estimation usually involves triangulation from two translation directions obtained from two different reference images\cite{hartley2003multiple, zhou2020learn}. However, the proposed method incorporates the local planar motion property not only into the essential matrix estimation, but also into the triangulation process in a creative way, thereby further reducing the number of required features, as shown in Fig. \ref{fig.overall}. To the best of our knowledge, this is the first minimal solution that models the local planar motion characteristics into triangulation. In addition, we propose a multiple checking procedure to ensure the robustness of the proposed visual localization method.

\section{Overview}
\begin{figure*}[tbp]
    \centering
    \includegraphics[width=1.0\textwidth]{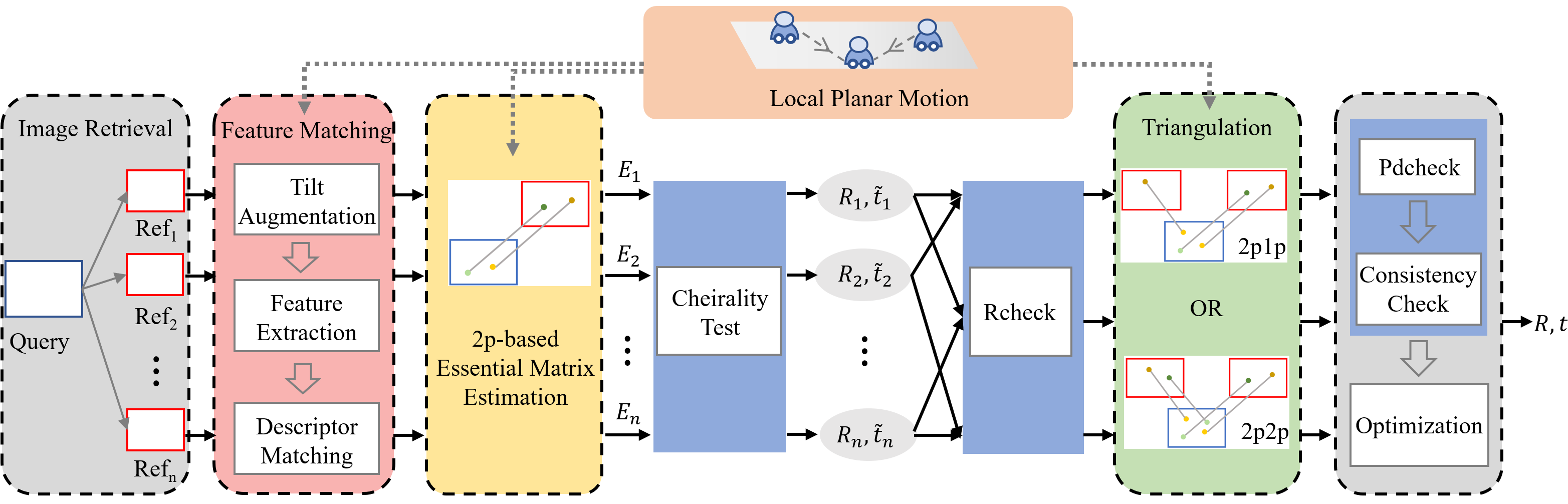}
    \caption{The pipeline of the proposed 3D model-free visual localization system. The local planar motion property is introduced into feature matching, essential matrix estimation and triangulation stages, which improve overall performance and robustness of the proposed visual localization system. A multiple checking procedure is also introduced to enhance the correctness of the pose estimation.}
    \label{fig.pipeline}
\end{figure*}

The pipeline of the proposed 3D model-free visual localization system under local planar motion is depicted in Fig. \ref{fig.pipeline}. Initially, image retrieval is conducted to identify a set of reference images that correspond to the same location as the query image in the database. Subsequently, augmentation-based feature matching is performed to establish correspondences between the query image and each retrieved reference image (\textit{c.f.} Sec. \ref{sec.feature}). During this stage, the local planar motion property (\textit{c.f.} Sec. \ref{sec.planar}) is exploited to relax the affine distortion formulation, thereby achieving affine invariant feature matching. This approach significantly enhances the robustness of feature matching against significant viewpoint variations, resulting in a considerable increase in the number of inlier matches.

Using the established correspondences, the essential matrix is computed for each pair of query and reference images, encoding the relative pose transformation (\textit{c.f.} Sec. \ref{sec.2p}). The relative poses that pass the cheirality test and Rcheck are then fed into the triangulation stage to estimate the absolute pose (\textit{c.f.} Sec. \ref{sec.2p2p} and Sec. \ref{sec.2p1p}). Following the Pdcheck and consistency check, we perform 6DoF nonlinear optimization to refine the estimated localization result using identified inliers. To enhance the robustness, the local planar motion property is incorporated into both the essential matrix estimation and triangulation stages. This integration aims to reduce the minimum number of correspondences required for accurate relative and absolute pose estimation. By leveraging the inherent local planar motion property, both feature matching and pose estimation become more resilient to noise and outliers, thereby improving overall performance and robustness of the proposed 3D model-free visual localization system.

\subsection{Local Planar Motion Property}\label{sec.planar}
\begin{figure}[tbp]
    \centering
    \includegraphics[width=0.3\textwidth]{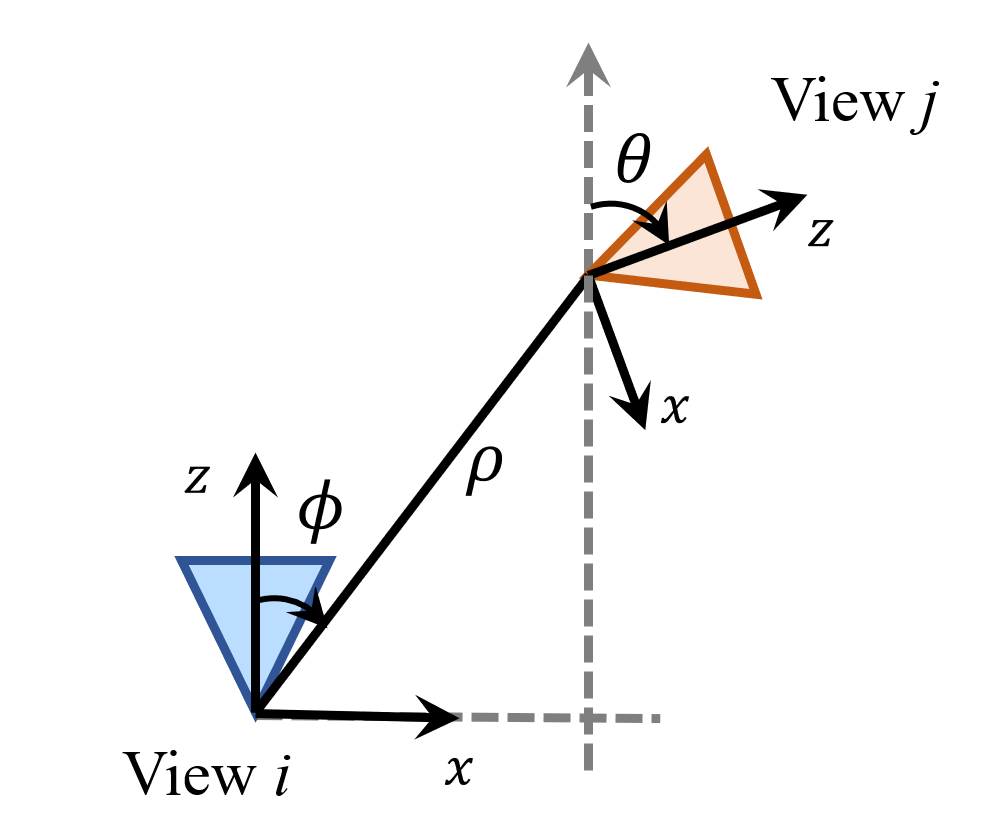}
    \caption{The illustration of local planar motion between query and reference views. There are three unknowns: yaw angle $\theta$, translation direction $\phi$ and scale $\rho$ in absolute pose estimation problem. }
    \label{fig.coordinate}
\end{figure}

Ground mobile robots basically have an inherent motion property, \textit{i.e.}, a local planar motion property. As shown in Fig. \ref{fig.coordinate} (a), there are only a $y$-axis rotation and 2D translation between query and reference camera views. Then the camera pose transformation from query view $i$ to reference view $j$ can be represented as
\begin{equation}\label{eq.Rt}
\mathbf{R}=\begin{bmatrix}
                \cos\theta& 0 & -\sin\theta \\
               0  & 1 & 0 \\
               \sin\theta & 0 & \cos\theta
             \end{bmatrix}, \mathbf{t}=-\mathbf{R}\rho{\begin{bmatrix}
      \sin\phi \\
      0 \\
      \cos\phi
    \end{bmatrix}}
\end{equation}
where $\theta$ denotes the yaw angle around axis $y$, and $\rho$, $\phi$ present the scale and direction of 2D translation on the $xz$ polar coordinate. Then the essential matrix $\mathbf{E}=[\mathbf{t}]_{\times}\mathbf{R}$ under local planar motion is reformulated as:
\begin{equation}\label{eq.E}
\mathbf{E}=\rho\begin{bmatrix}
                 0 & \cos(\theta-\phi) & 0 \\
                 -\cos\phi & 0 & \sin\phi \\
                 0 & \sin(\theta-\phi) & 0
               \end{bmatrix}
\end{equation}
where $[\cdot]_{\times}$ denotes the skew symmetry matrix.
Therefore, it is imperative to construct a dedicated visual localization system that reformulates feature matching and pose estimation based on the local planar motion property, rather than persisting with a general system designed for the 6 DoF problem.

\subsection{Augmentation-based Feature Matching}\label{sec.feature}
For robust feature matching, we adopt TiltR2D2 \cite{jiao2022leveraging} in our previous work to perform affine-invariant feature matching. The local planar motion property is formulated into projective deformation between query and reference images, leading to a simplified affine distortion model. Specifically, there are only a transition tilt which represents a longitudinal stretch of the image and a scale transformation between two views under local planar motion. Then we design a tilt augmentation strategy to achieve tilt-invariant which can be then integrated with the existing scale-invariant descriptors to achieve full affine invariance. For details about TiltR2D2, please refer to \cite{jiao2022leveraging}.

\section{Minimal Solutions from Essential Matrix}\label{sec.solution}
For robust pose estimation, two minimal solutions from essential matrix are derived to perform robust 3D model-free visual localization. The key to reduce the number of feature correspondences for pose estimation is to leverage local planar motion property of wheeled mobile robots. Then we derive one minimal solution namely 2p2p which uses only 2 point correspondences for essential matrix estimation in Sec. \ref{sec.2p} and another 2 correspondences for absolute pose triangulation in Sec. \ref{sec.2p2p}. And the minimal solution namely 2p1p is derived in Sec. \ref{sec.2p1p}, which only requires one additional correspondence for absolute scale estimation.

\subsection{2p-based Essential Matrix Estimation}\label{sec.2p}
Given one 2D-2D feature correspondence between query view and reference view expressed as $p_i$ in query view and $p_j$ in reference view, the epipolar constraint is given as follows:
\begin{equation}\label{eq.epipolar}
  p_j^\mathrm{T}\mathbf{E}p_i=0
\end{equation}
where $p_i=[\tilde{u}_{i},\tilde{v}_{i},1]^\mathrm{T}=K^{-1}[{u}_{i},{v}_{i},1]^\mathrm{T}$ and $p_j=[\tilde{u}_{j},\tilde{v}_{j},1]^\mathrm{T}=K^{-1}[{u}_{j},{v}_{j},1]^\mathrm{T}$ are the normalized homogeneous coordinates of a feature point in view $i$ and $j$, respectively. $K$ is the calibrated intrinsic matrix.

By substituting (\ref{eq.E}) into (\ref{eq.epipolar}), the epipolar constraint under local planar motion can be reformulated as:
\begin{equation}\label{eq.Epipoloar1}
  \tilde{v}_{i}\sin(\theta-\phi)+\tilde{v}_{i}\tilde{u}_{j}\cos(\theta-\phi)+\tilde{v}_{j}\sin\phi-\tilde{u}_{i}\tilde{v}_{j}\cos\phi=0
\end{equation}

Given another 2D-2D feature correspondence denoted as $p_{i_2}$ in query view and $p_{j_2}$ in reference view, another constraint can be obtained as:
\begin{equation}\label{eq.Epipoloar2}
  \tilde{v}_{i_2}\sin(\theta-\phi)+\tilde{v}_{i_2}\tilde{u}_{j_2}\cos(\theta-\phi)+\tilde{v}_{j_2}\sin\phi-\tilde{u}_{i_2}\tilde{v}_{j_2}\cos\phi=0
\end{equation}

For each two correspondences, the combination of (\ref{eq.Epipoloar1})-(\ref{eq.Epipoloar2}) can be expressed as
\begin{equation}\label{eq.ax}
\mathbf{A}\mathbf{x}=\mathbf{0}
\end{equation}
where $\mathbf{x}=[\sin(\theta-\phi),\cos(\theta-\phi),\sin\phi,\cos\phi]^\mathrm{T}$. To facilitate the description of the following derivation, we denote
\begin{equation}\label{eq.ax0}
\left\{
\begin{aligned}
    &x_1 \triangleq \sin(\theta-\phi), &&x_2 \triangleq \cos(\theta-\phi) \\
    &x_3 \triangleq \sin\phi, &&x_4 \triangleq \cos\phi
  \end{aligned}
\right.
\end{equation}
Then (\ref{eq.ax}) can be reformulated as
\begin{equation}\label{eq.ax12}
a_ix_1+b_ix_2+c_ix_3+d_ix_4=0, i=1,2
\end{equation}
where coefficients $a_i$, $b_i$, $c_i$ and $d_i$ denote the problem coefficients in (\ref{eq.Epipoloar1})-(\ref{eq.Epipoloar2}).

According to (\ref{eq.ax12}), $\mathbf{x}$ should lie in the null space of $\mathbf{A}$. Then we can obtain the general solution of $\mathbf{x}$ as
\begin{equation}\label{eq.ax3}
\mathbf{x} = \lambda_1 \mathbf{X}_1 + \lambda_2 \mathbf{X}_2
\end{equation}
where $\mathbf{X}_1$ and $\mathbf{X}_1$ are the null vectors of $\mathbf{A}$.

There are implicit constraints of trigonometric functions between the entries of $\mathbf{x}$ as follows:
\begin{equation}\label{eq.ax4}
\left\{
\begin{aligned}
    &x_1^2+x_2^2=1 \\
    &x_3^2+x_4^2=1
\end{aligned}
\right.
\end{equation}
By substituting the general solution (\ref{eq.ax3}) into (\ref{eq.ax4}), the special solution of $\mathbf{x}$ can be solved. Once $\mathbf{x}$ has been obtained, the angles of $\phi$ and $\theta$ are
\begin{equation}\label{eq.ax5}
  \left\{
\begin{aligned}
      &\phi = \arctan2(x_3,x_4), \\
      &\theta = \arctan2(x_1,x_2)+\phi
\end{aligned}
\right.
\end{equation}

Then the essential matrix can be obtained from (\ref{eq.E}).

\subsection{2p2p-based Absolute Pose Estimation}\label{sec.2p2p}
For each query image, top-$n$ reference images can be picked by performing image retrieval method such as NetVLAD \cite{arandjelovic2016netvlad}. Therefore, there are total $n$ image pairs as shown in Fig. \ref{fig.pipeline}. For each image pair, we compute the essential matrix by selecting any 2 correspondences as derived in Sec. \ref{sec.2p}. Next, we extract the four relative poses $(\mathbf{R},\mathbf{\tilde{t}})$, $(\mathbf{R},-\mathbf{\tilde{t}})$, $(\mathbf{R}',\mathbf{\tilde{t}})$,$(\mathbf{R}',-\mathbf{\tilde{t}})$ from essential matrix, where $\mathbf{\tilde{t}}$ is the normalized relative translation vector and $\|\mathbf{\tilde{t}}\|=1$, and $\mathbf{R}$ and $\mathbf{R}'$ are related by a 180\degree \quad
rotation around the baseline \cite{hartley2003multiple}. Traditionally, we perform a \textbf{cheirality test} using the feature correspondences to select the correct relative pose among the four candidates as in \cite{hartley2003multiple}.

After extracting the relative pose of each image pair, we obtain the multiple directions between the query camera center and the top-$n$ reference camera centers $\mathbf{\tilde{t}}_1,...,\mathbf{\tilde{t}}_n$. Then the absolute position of the query image can be uniquely determined by performing triangulation from each two directions. Thus the absolute pose of the query image can be solved which utilizes 2 point correspondences from 2 reference images, respectively. The solution is denoted as 2p2p. To deal with outliers in correspondences, we propose a robust estimation framework by performing a multiple checking procedure to speed up the outlier rejection and add multiple checks for filtering the optimal solution, as shown in Fig. \ref{fig.pipeline}.

\textbf{Rcheck}: We can perform a Rotation Check (Rcheck) before triangulation when selecting two relative poses  $(\mathbf{R},\mathbf{\tilde{t}})$ and $(\mathbf{R}_2,\mathbf{\tilde{t}}_2)$ as follows. According to the transformation matrix representation between the query view $i$ and the two selected reference views $j$ and $j_2$, we have
\begin{equation}\label{eq.rcheck}
\begin{aligned}
  \mathbf{T}_{jj_2} &= \mathbf{T}_{ji}\mathbf{T}_{j_2i}^{-1}\\
  \textit{i.e.}, \begin{bmatrix}
         \mathbf{R}_{12} & \mathbf{t}_{12} \\
         \mathbf{0} & 1
       \end{bmatrix} &=
       \begin{bmatrix}
         \mathbf{R} & \rho\mathbf{\tilde{t}} \\
         \mathbf{0} & 1
       \end{bmatrix}
       \begin{bmatrix}
         \mathbf{R}_{2} & -\rho_2\mathbf{R}_{2}^\mathrm{T}\mathbf{\tilde{t}}_{2} \\
         \mathbf{0} & 1
       \end{bmatrix}\\
       &=
       \begin{bmatrix}
         \mathbf{R}\mathbf{R}_{2}^\mathrm{T} & -\rho_2\mathbf{R}\mathbf{R}_{2}^\mathrm{T}\mathbf{\tilde{t}}_{2}+\rho\mathbf{\tilde{t}} \\
         \mathbf{0} & 1
       \end{bmatrix}
\end{aligned}
\end{equation}

From the rotation part in (\ref{eq.rcheck}), we can find that $\mathbf{R}_{12} = \mathbf{R}\mathbf{R}_{2}^\mathrm{T}$ must be satisfied for two inlier relative poses. Note that $\mathbf{T}_{jj_2}$ is the known transformation matrix between two reference views which can be retrieved from pre-built database. Subsequent triangulation is performed only if the selected relative position can pass the rotation check. Considering measurement noise and calibration error, we set a threshold $e_{th}$ to perform a relaxed Rcheck. Specifically, we compute the error between the two rotation matrix as $e=\arccos(0.5Tr(\mathbf{R}_{12}(\mathbf{R}\mathbf{R}_{2}^\mathrm{T})^\mathrm{T})-0.5)$ in degree. The Rcheck is considered passed if $e \leq e_{th}$.

\textbf{Triangulation and Pdcheck}: The absolute position of the query view can be solved through triangulation by solving the translation part in (\ref{eq.rcheck}) as follows. Denoting $\mathbf{s}=[\rho,\rho_2]^\mathrm{T}$, the translation part in (\ref{eq.rcheck}) can be expressed as
\begin{equation}\label{eq.ax=b}
  \mathbf{C}\mathbf{s}=\mathbf{b}
\end{equation}
where $\mathbf{C}=[\mathbf{\tilde{t}},-\mathbf{R}\mathbf{R}_{2}^\mathrm{T}\mathbf{\tilde{t}}_{2}]$, $\mathbf{b}=\mathbf{t}_{12}$. Then we find least-squared solution of $\mathbf{s}$ by
\begin{equation}\label{eq.tls}
  \mathbf{s}=(\mathbf{C}^\mathrm{T}\mathbf{C})^{-1}\mathbf{C}^\mathrm{T}\mathbf{b}
\end{equation}
Note that the above solution procedure holds only when the $\mathbf{C}^\mathrm{T}\mathbf{C}$ is invertible. If $Det(\mathbf{C}^\mathrm{T}\mathbf{C}) = 0$, the regular term needs to be added to make it invertible, as follows
\begin{equation}\label{eq.tls}
  \mathbf{s}=(\mathbf{C}^\mathrm{T}\mathbf{C}+\mathbf{I})^{-1}\mathbf{C}^\mathrm{T}\mathbf{b}
\end{equation}
where $\mathbf{I}$ is $2\times2$ identity matrix.
After triangulation, there is a Positive Depth Check (Pdcheck) which requires translation scales to be positive as $\rho>0, \rho_2>0$.


\textbf{Consistency Check}: With the estimated scale, we can obtain the absolute pose of query view through the known pose of reference view in database. Then we can measure the consistency between the relative translation obtained from the estimated absolute pose and the retrieved relative translation from the essential matrix. Specifically, denoting the camera center of query view as $c_i$ and reference view as $c_j$, the relative translation between query view and reference view is $\mathbf{\hat{t}}=\mathbf{R}_j^\mathrm{T}(c_i-c_j)$. And $\mathbf{R}_j$ is the rotation which rotates the vector in reference view to the global coordinate system. The direction of $\mathbf{\hat{t}}$ should be consistent with the direction of $\mathbf{\tilde{t}}$ recovered from the essential matrix. For quantification, we compute the angle $\alpha=\arccos(\mathbf{\tilde{t}}^\mathrm{T}\mathbf{\hat{t}}/\|\mathbf{\hat{t}}\|)$. If the angle between the two predicted translation directions is less than a given threshold $\alpha_{th}$, the consistency check is considered passed.

\textbf{Nonlinear Optimization}: Finally, we perform 6DoF nonlinear optimization using the identified 2D-2D inliers (and 2D-3D inliers if there is depth information in database) to optimize the absolute pose corresponding to the largest number of inliers in RANSAC. The nonlinear optimization can correct for noise in other DoF caused by occasional uneven ground.

\begin{figure}[tbp]
    \centering
    \includegraphics[width=0.5\textwidth]{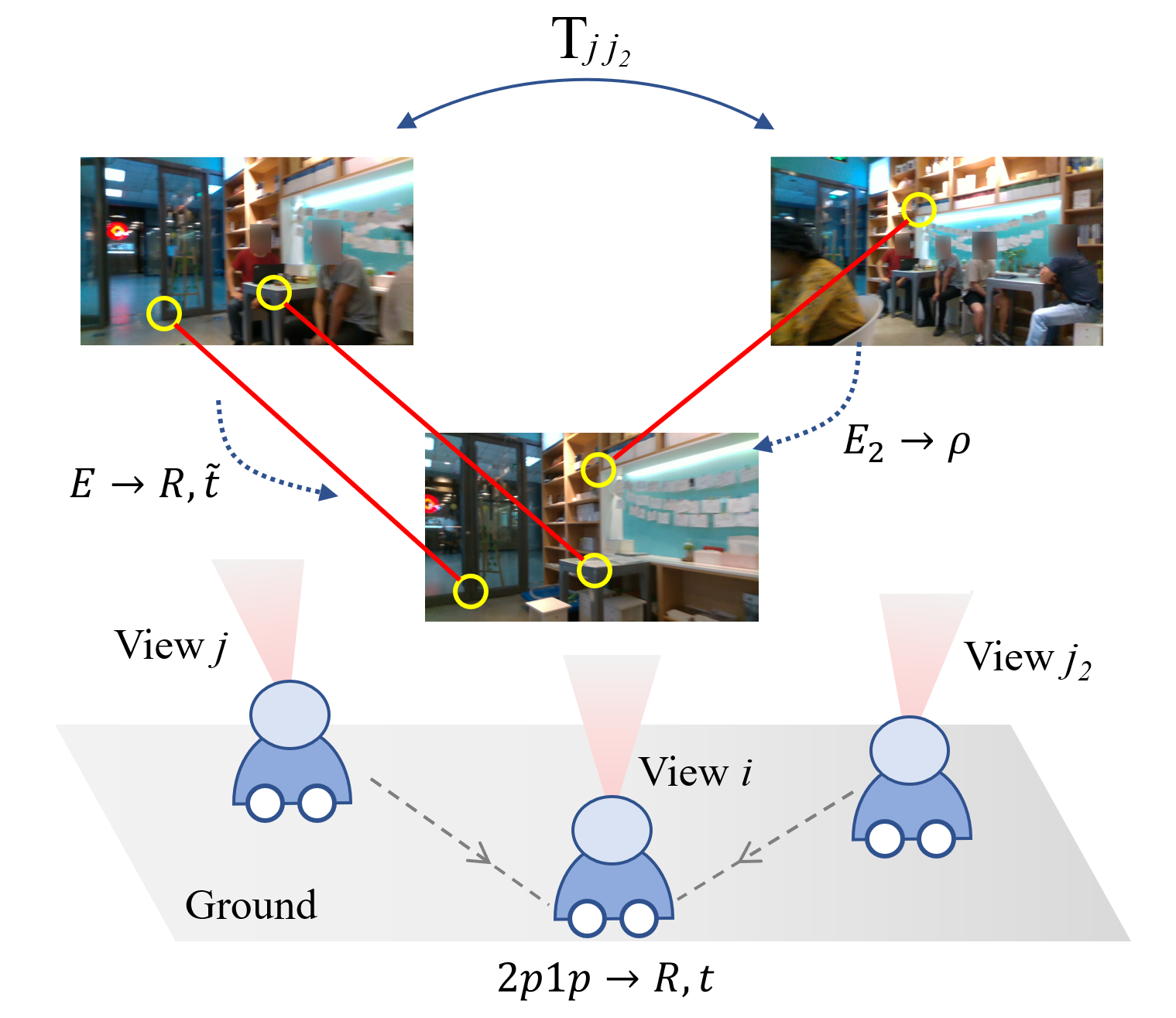}
    \caption{Illustration of proposed 2p1p minimal solution for absolute pose estimation. By exploiting the local planar motion property, this solution requires only 2 points for essential matrix estimation and 1 point for triangulation, enhancing the robustness of the 3D model-free visual localization system.}
    \label{fig.overall}
\end{figure}

\subsection{2p1p-based Absolute Pose Estimation}\label{sec.2p1p}
From the derivation in Sec. \ref{sec.2p}, we can find that essential matrix can be solved using 2 point correspondences between the query view and one reference view. Then only the absolute translation scale is unknown which needs to be solved with information provided by another reference view. As one point correspondence can provide one constraint based on epipolar constraint as in (\ref{eq.epipolar}), the minimal solution for absolute pose estimation from essential matrix could be 2p1p, which means only one point correspondence from another reference view is required to solve the unknown scale, as shown in Fig. \ref{fig.overall}. The details in derivation of 2p1p-based minimal solution is shown as follows.

As in Sec. \ref{sec.2p}, we first solve the essential matrix using 2 point correspondences in reference view $j$. Then with the known transformation matrix $\mathbf{T}_{j_2j}$ between reference view $j$ and $j_2$ retrieved from the database, we can express the $\mathbf{T}_{j_2i}$ using the unknown absolute scale $\rho$ in $\mathbf{T}_{ji}$ as follows:
\begin{equation}\label{eq.ij2}
\begin{aligned}
  \mathbf{T}_{j_2i} &= \mathbf{T}_{j_2j}\mathbf{T}_{ji}\\
  &=\begin{bmatrix}
         \mathbf{R}_{21} & \mathbf{t}_{21} \\
         \mathbf{0} & 1
       \end{bmatrix}
       \begin{bmatrix}
         \mathbf{R} & \rho\mathbf{\tilde{t}} \\
         \mathbf{0} & 1
       \end{bmatrix}
       \\
       &=
       \begin{bmatrix}
         \mathbf{R}_{21}\mathbf{R} & \rho\mathbf{R}_{21}\mathbf{\tilde{t}}+\mathbf{t}_{21} \\
         \mathbf{0} & 1
       \end{bmatrix}
\end{aligned}
\end{equation}

Then the essential matrix between query view $i$ and reference view $j_2$ can be expressed with the unknown scale $\rho$ as $\mathbf{E}_2=[\mathbf{t}_{j_2i}]_{\times}\mathbf{R}_{j_2i}=[\rho\mathbf{R}_{21}\mathbf{\tilde{t}}+\mathbf{t}_{21}]_{\times}\mathbf{R}_{21}\mathbf{R}$.
Given one 2D-2D feature correspondence denoted as $p_{i_3}$ in query view and $p_{j_3}$ in reference view $j_2$, we can obtain one constraint about $\rho$ using the epipolar constraint as:
\begin{equation}\label{eq.j3}
  p_{j_3}^\mathrm{T}\mathbf{E}_2p_{i_3}=0
\end{equation}
\textit{i.e.},
\begin{equation}\label{eq.j32}
  p_{j_3}^\mathrm{T} [\rho\mathbf{R}_{21}\mathbf{\tilde{t}}+\mathbf{t}_{21}]_{\times}\mathbf{R}_{21}\mathbf{R} p_{i_3}=0
\end{equation}
By solving (\ref{eq.j32}), we can uniquely determine the unknown scale $\rho$. We denote this minimal solution as 2p1p, which can be integrated into the RANSAC framework for robust pose estimation. Furthermore, we can apply the multiple checking procedure proposed in Sec. \ref{sec.2p2p} to enhance the robustness of our method.

\begin{figure*}[tbp]
    \centering
    \includegraphics[width=1.0\textwidth]{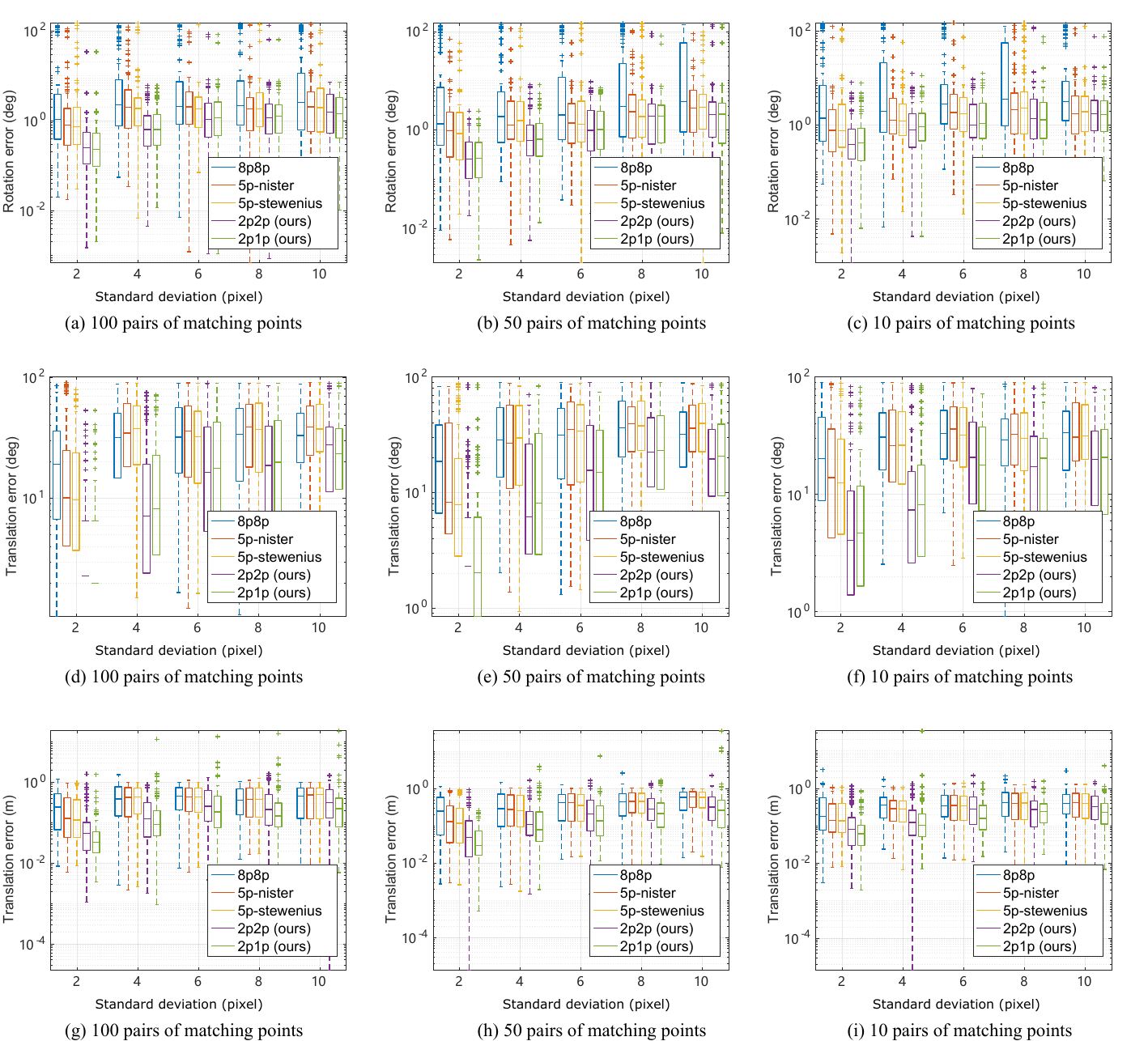}
    \caption{Comparison of pose estimation accuracy with increasing 2D noise.}
  \label{fig.2dexp}
\end{figure*}

\section{Experimental Results}
To validate the performance of the proposed minimal solutions and the 3D model-free visual localization system, we conduct simulation experiments using synthetic data and real world experiments on two indoor datasets containing lifelong changes. Specifically, we perform simulation experiments to test (i) the accuracy of the proposed minimal solutions, and (ii) the robustness of the proposed multiple-checking based framework against increasing outlier rate of correspondences, and (iii) the efficiency of the proposed minimal solutions by counting the computation time. In real world experiments, we first (iv) conduct ablation studies to show the effectiveness of each stage in the system, and (v) compare the success rate to show the robustness of the whole localization system in changing environments, and (vi) demonstrate the localization performance by comparing the whole trajectory, and finally (vii) show several special cases to illustrate the effectiveness of the proposed system.

For comparison, we select the most classical 8-point solution \cite{fischler2014readings} and two variations of 5-point minimal solution under 6DoF problem formulation\cite{nister2004efficient, stewenius2006recent} to compute the essential matrix. we adopt the implementation of above algorithms in OpenGV\cite{kneip2014opengv}. To achieve absolute pose estimation, we perform the general triangulation for these solutions and the obtained methods are denoted as \emph{8p8p}, \emph{5p5p-stewenius} and \emph{5p5p-nister}, respectively. We implement the proposed \emph{2p2p} and \emph{2p1p} minimal solutions and the multiple checking framework in Matlab. The $e_{th}$ and $\alpha_{th}$ in Rcheck and Consistency check are empirically set to 2 \degree. All experiments are performed on a desktop with CPU Intel i9-12900H 2.50GHz and 16G RAM.

\subsection{Simulation Experiments}\label{exp.sim}
To compare the theoretic performance of different absolute pose estimation algorithms, we generate test data in a synthetic world as in \cite{jiao2020robust}. We first randomly generate 3D points in a cube $[-10,10]^{3}$ as map points in the world. Then we project a number of co-visible points to query view and multiple reference views to obtain 2D-2D feature correspondences. All views are captured from a virtual camera with random poses sampled under local planar motion property. Specifically, the translation of the virtual camera is varying in range of $[-5,5]$ along $x$-axis and $z$-axis, and the rotation is varying in range of $[-\pi,\pi]$ around $y$-axis. The focal length of the virtual camera is fixed as 800 with the resolution of 1280$\times$1080 and the principal point as $(640,540)$. For quantitative evaluation, we measure the error between the ground truth of the pose of query view $[\mathbf{R}_{gt}|\mathbf{t}_{gt}]$ and the estimated pose $[\mathbf{R}|\mathbf{t}]$. Similar in \cite{sattler2018benchmarking}, we calculate the rotation error as $\arccos (0.5Tr(\mathbf{R}\mathbf{R}_{gt}^T)-0.5)$ in degree, and the translation error as $|\mathbf{t}-\mathbf{t}_{gt}|$ in meter. And the relative translation vector error is computed as $\arccos(\mathbf{\tilde{t}}^\mathrm{T}\mathbf{{t}}_{gt}/\|\mathbf{{t}}_{gt}\|)$, where $\mathbf{\tilde{t}}$ is the normalized translation direction of $\mathbf{t}$.

\textbf{Accuracy}:
We conduct accuracy evaluation experiment to verify the performance of the proposed pose estimation method under varying 2D noise levels and numbers of feature matches. Specifically, we add Gaussian nose with zero mean and increasing standard deviation from 2 to 10 pixels to the feature points of 2D-2D matching, forming eleven levels of noise. At the same time, we design three experiment groups with varying number of total feature matches: 100, 50, and 10. We generate 100 experimental tests for each noise level and perform 100 iterations of RANSAC for each method. The mean error of rotation estimation, translation direction and absolute translation estimation are shown in Fig. \ref{fig.2dexp}.

The results of rotation and translation direction estimation error in Fig. \ref{fig.2dexp} (a)-(f) demonstrate the accuracy of different essential matrix estimation solutions. The proposed 2p-based method exhibits high accuracy under different levels of noise and number of feature matches, which is shown in results of 2p2p and 2p1p solutions. The performance of absolute pose estimation can be evaluated from the error of absolute translation estimation shown in Fig. \ref{fig.2dexp} (g)-(i). The proposed solutions outperform the compared methods, which is consistent with the results of essential matrix estimation. Particularly, even under high levels of noise and low numbers of correspondences, the proposed solutions are still able to achieve high accuracy in pose estimation. The accuracy evaluation experiment validates the effectiveness of leveraging local planar motion property into pose estimation. This is because less noise is introduced into pose estimation, leading to higher accuracy. In addition, the translation error of the proposed 2p1p minimal solution is smaller than the 2p2p solution, which further verifies the importance of formulating the motion property into triangulation.

\begin{figure*}[tbp]
  \centering

  \subfigure[100 pairs of matching points]{
            \includegraphics[width=0.3\textwidth]{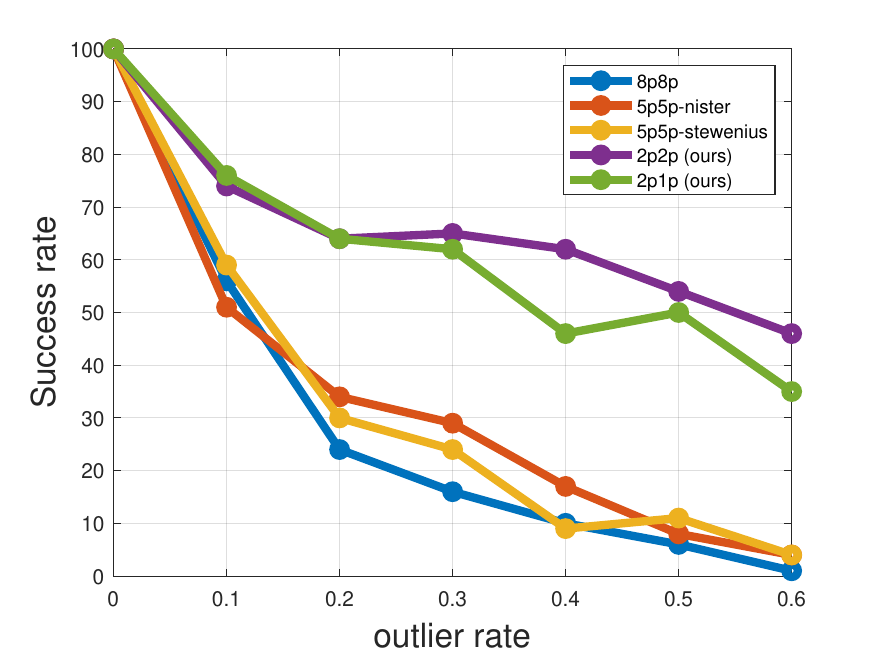}
        }
  \subfigure[50 pairs of matching points]{
            \includegraphics[width=0.3\textwidth]{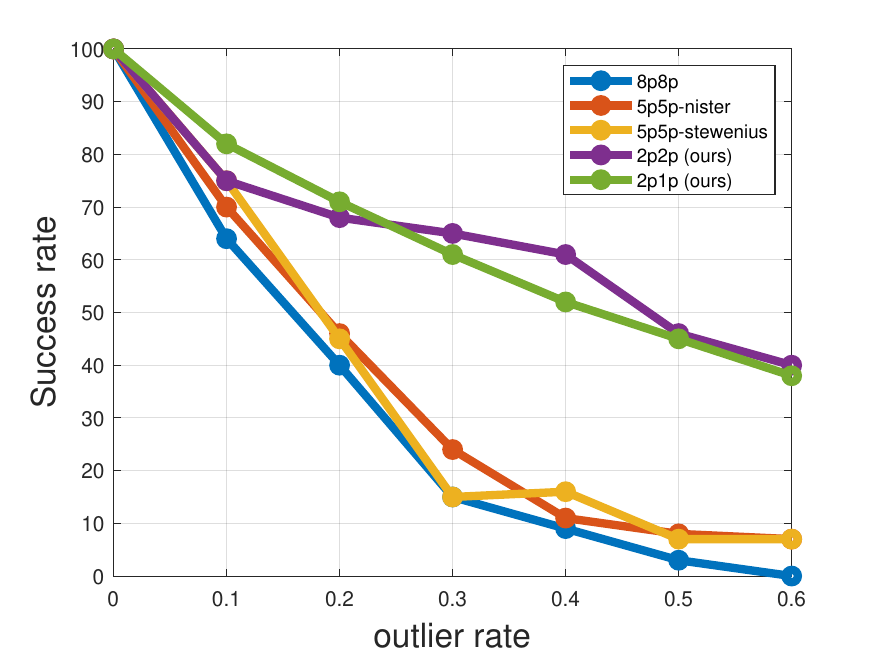}
        }
  \subfigure[20 pairs of matching points]{
            \includegraphics[width=0.3\textwidth]{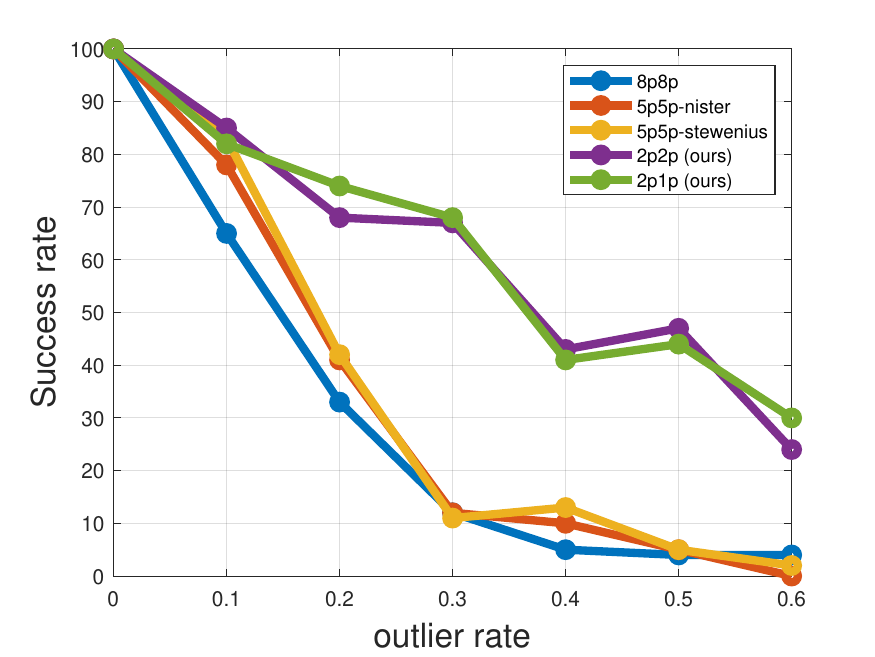}
        }

  \caption{Comparison in robustness of pose estimation algorithms with increasing outlier rate.}
  \label{fig.outexp}
\end{figure*}

\textbf{Robustness}:
We conduct a robustness simulation experiment by adding increasing rates of outliers from 0\% to 60\% to the feature matches. Outliers are generated by projecting points using incorrect poses. The success rate of pose estimation is counted for each method to demonstrate robustness. There are three experimental groups according to the number of total feature matches: 100, 50, and 20. Note that the minimum number of feature matches is changed to 20, which is different from 10 in the accuracy experiment, as there are not enough inliers to perform pose estimation for the 8p8p method when the outlier rate increases to 60\%. We consider the localization to be successful when the translation error is less than 0.1 m and the rotation error is less than 1 \degree. The comparison of the success rate of different methods is shown in Fig. \ref{fig.outexp}.

Based on the experimental results, we can see that the performance of all methods gradually decreases as the outlier rate increases. However, the proposed methods (2p1p and 2p2p) outperform the comparison methods in all three experimental tests. The advantage of the proposed methods is more obvious with high outlier rate conditions, where all comparison methods failed to localize (e.g., outlier rate of 60\%), while the proposed methods can still maintain a success rate of 30\%-50\%. And when the outlier rate is 60\%, a comparison of the results between the proposed 2p1p and 2p2p methods under three different feature counts reveals that, as the number of features decreases, the success rate of the 2p1p method gradually surpasses that of the 2p2p method. This is because the 2p1p method requires fewer minimum feature matches for calculation, and its advantage becomes more pronounced when there is a high proportion of outliers with a low number of total features. Experimental results show that the proposed methods are more robust to outliers and can better handle the case of small number of feature matches, of which the situation is common in environments with lifelong changing variations in real-world robotic applications.

\begin{table}[tbp]
	\caption{Computation time evaluation (ms)}
	\begin{center}
		\resizebox{0.45\textwidth}{!}{
			\begin{threeparttable}
				\begin{tabular}{llcccccc}
					\Xhline{1pt}
					\multirow{3}{*}{} & Points      & 10                     & 50                     & 100                   & 200               & 500     &                     \\
				
					\Xhline{1pt}
					\multirow{5}{*}{}&2p2p             & 1.394     &   1.673   &  1.777   & 2.354  & 5.638 &  \\
&2p1p            &  0.532      &     0.557   & 0.613  &  1.046 &   1.983 & \\
										\Xhline{1pt}

				\end{tabular}
				
		\end{threeparttable}}
	\end{center}
	\label{table.time}
\end{table}

\textbf{Efficiency}: The real-time performance is important for visual localization algorithms. We vary the number of point correspondences in the scene to demonstrate the efficiency of the proposed methods. In the test data, we add the Gaussian noise with zero mean and 5 pixel standard deviation and generate 10\% outliers to simulate the real condition. We run the whole multiple checking procedure for 100 times to count the average execution time. The computation time evaluation results with increasing number of points are shown in Table. \ref{table.time}. For the sake of fair comparison, the performance of compared methods implemented in C++ is not presented. It is important to note that our algorithms are currently implemented in MATLAB, which can be more efficient in C++. The results demonstrate that the proposed methods can achieve real-time performance, which is suitable for robotics tasks. In addition, as shown in Table. \ref{table.time}, by introducing motion constraints to the triangulation stage, a significant reduction in computational complexity is achieved, resulting in a significantly faster computing speed for 2p1p compared to 2p2p.

\subsection{Experiments on OpenLORIS}
In real-world experiments evaluation, we first perform thorough evaluations on a public indoor dataset: OpenLORIS-Scene dataset\cite{shi2020we}. OpenLORIS-Scene is a multi-modal dataset containing RGB images, depth images and LiDAR data collected by a wheeled robot. The scenes in the dataset include common indoor environments such as offices, cafes, corridors and bedrooms. There are challenging lifelong variations including viewpoints, illuminations, dynamic objects changes and frequent human activities, which are generic and representative for localization algorithms validation.

\textbf{Experimental Settings}: We test the localization performance on 4 different scenes, including \textit{home}, \textit{corridor}, \textit{office} and \textit{cafe}. There are different number of sequences in each scene which are collected in different time periods. Only RGB images in the dataset are used in the experiments for all methods. There are four steps for evaluation: (1) we perform NetVLAD algorithm\cite{arandjelovic2016netvlad} to retrieve reference images from database for each query image. (2) for feature matching between each retrieved image pairs, we use R2D2 algorithm\cite{revaud2019r2d2} as a representative method in general motion, serving as a comparative method with the proposed TiltR2D2 in local planar motion. (3) using the known absolute poses of the retrieved reference images and 2D-2D feature matches, we estimate the absolute pose of the query image using different methods, respectively. (4) for quantitative evaluation, we measure the translation and rotation error between the ground truth pose and estimated poses and count the success rate under different error thresholds. 100 iterations are performed for each solution in RANSAC and nonlinear optimization is applied for further refinement.

\begin{figure}[tbp]
    \centering
    \includegraphics[width=0.5\textwidth]{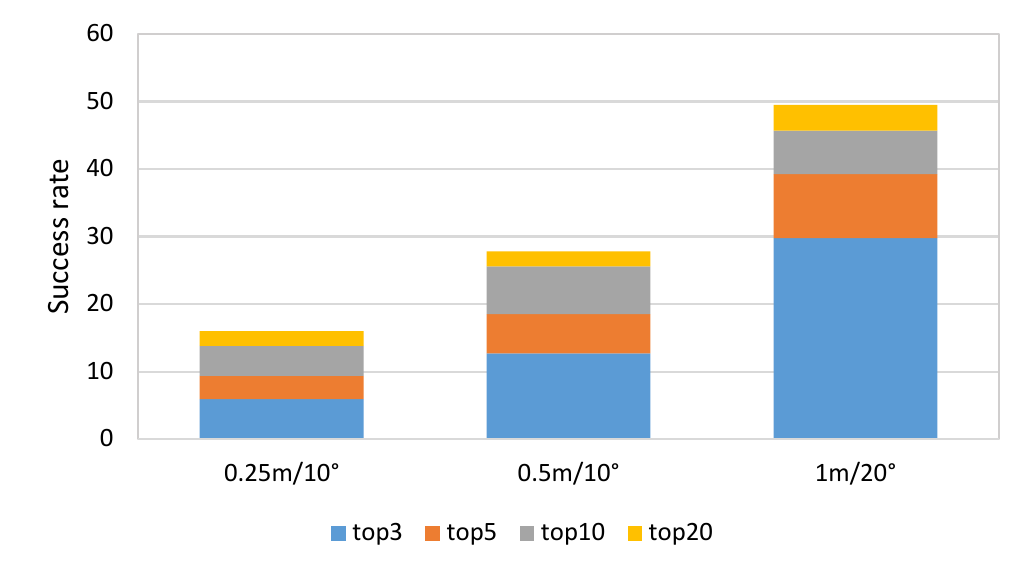}
    \caption{The result of varying the number of reference images in the image retrieval stage on the localization performance.}
    \label{fig.netvlad}
\end{figure}

\textbf{Ablation Study}: We conduct several experiments to validate the effectiveness of our design stages in the proposed 3D model-free localization system with success rate comparison in \textit{home4} session. We first evaluate the impact of varying the number of reference images in the image retrieval stage on the localization performance. The result, as depicted in Fig. \ref{fig.netvlad}, demonstrates that an elevated number of reference images positively influences the localization success rate by facilitating a linear growth in the quantity of feature matches available. However, when the number of images exceeds 5, the performance improvement noticeably slows down. Considering that extracting additional features also increases the computational burden for subsequent pose estimation, we uniformly choose to extract the top 5 reference images in the subsequent experiments.

\begin{table}[tbp]
	\caption{Ablation studies: success rate comparison on \textit{home4}.}
	\begin{center}
		\resizebox{0.5\textwidth}{!}{
			\begin{threeparttable}
				\begin{tabular}{llcccc}
					\Xhline{1pt}
					        & method                     & 0.25m/10\degree                     & 0.5m/10\degree                   & 1.0m/20\degree                              \\ \hline
					baseline                          &top5-R2D2-8p8p      &9.33          & 18.53         & 39.24           \\\hline
					+Tilt                      & top5-TiltR2D2-8p8p            & 10.94            & 19.87             & 40.07                      \\

					\rowcolor{mycyan} improvement     			 &                        & 1.61     & 1.34   & 0.83  \\
					+2p2p   & top5-R2D2-2p2p                          & 11.27     & 21.51   &42.79  \\
					\rowcolor{mycyan}improvement             &                        & 1.94  & 2.98 & 3.55   \\
					
					+2p1p     & top5-R2D2-2p1p                      & 11.46     & 21.69   & 43.66    \\
					\rowcolor{mycyan} improvement   &                & 2.13    & 3.16   & 4.42    \\
					
						+Tilt-2p2p          & top5-TiltR2D2-2p2p                        & 11.70 & 22.50 & 43.87   \\
						\rowcolor{mycyan} improvement             &                        & 2.37 & 3.97 & 4.63   \\
						+Tilt-2p1p             & top5-TiltR2D2-2p1p                       &12.96 & 23.16 & 44.34   \\
						\rowcolor{mycyan} improvement             &                        & 3.63 & 4.63 & 5.1   \\
						\Xhline{1pt}
				\end{tabular}
				\begin{tablenotes}
					\footnotesize
					\item[1] The improvement metrics in the table are all relative to the baseline.
				\end{tablenotes}
		\end{threeparttable}}
	\end{center}
	\label{table.ablation}
\end{table}

Then we quantitatively evaluate the impact of other stages in the localization pipeline, with the results shown in the Table. \ref{table.ablation}. We select the combination of R2D2 features and the classic 8p8p pose estimation as our baseline for comparison. Based on this baseline, we evaluate the feature matching and pose estimation methods separately, expressing the performance changes of each method compared to the baseline in the ``\textcolor{cyan}{improvement}'' row of the table, highlighted in cyan. The results indicate that the proposed feature matching method brings performance improvement to any pose estimation method, including 8p8p, 2p2p, and 2p1p. Due to the presence of object variations and occlusions caused by humans in the \textit{home4} scenario, the 2p1p method exhibits greater improvement compared to the 2p2p method. Furthermore, the proposed complete localization system under planar motion achieves optimal localization performance, as demonstrated by the ``Tilt-2p1p'' entry in the table, validating the effectiveness of the proposed localization system.

\begin{figure}[tbp]
    \centering
    \includegraphics[width=0.5\textwidth]{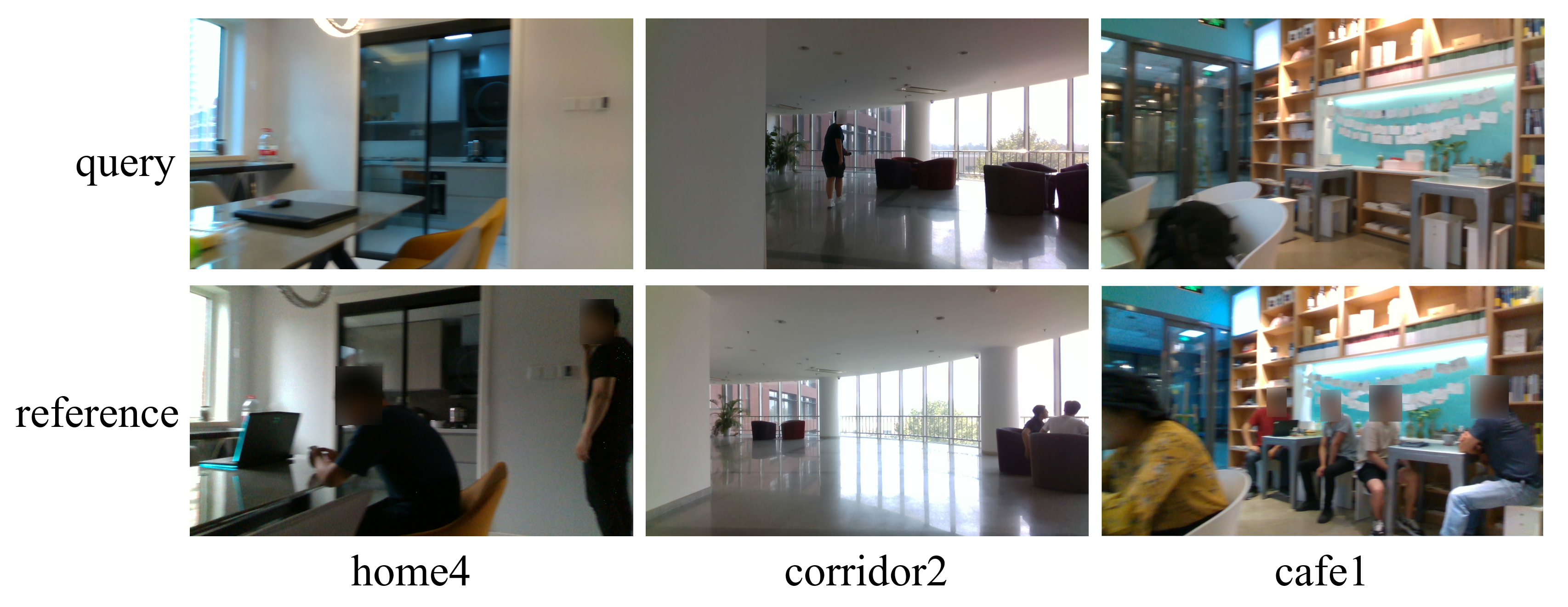}
    \caption{Localization examples extracted from \textit{home4}, \textit{corridor2} and \textit{cafe1}.}
    \label{fig.opencases}
\end{figure}

\textbf{Robustness Comparison}: The success rate comparison of different algorithms in all sessions of OpenLORIS-scene dataset is shown in Table. \ref{table.openloristable}. The results indicate that the proposed local planar motion aided localization system achieves the best performance in all scenes compared to the general 6DoF-based solvers, confirming the importance of leveraging motion property. In addition, comparing the results of Tilt-2p2p and Tilt-2p1p, we can find that the proposed Tilt-2p1p performs better in scenes where the localization success rate is less than 50\% (such as \textit{home4}, \textit{corridor2}, \textit{cafe1}). This is due to the fact that the challenges in these particular scenes are greater. For instance, frequent human activities in \textit{home} and \textit{cafe} scenes often result in dynamic occlusion and object movements, as shown in Fig. \ref{fig.opencases}. Additionally, significant illumination changes occur in the corridor scene due to the presence of a large French window. The aforementioned challenges directly result in a lower number of matched features extracted from these scenes, along with a high rate of outliers. In such circumstances, the proposed Tilt-2p1p's advantage of requiring fewer features for localization becomes more pronounced, which is consistent with the aforementioned analysis in simulation experiments. In scenes where the localization success rate is generally higher than 50\%, the advantage of the proposed Tilt-2p2p can be fully utilized thanks to the abundance of correct feature matches. Overall, the proposed methods outperform the comparative methods across the entire dataset, thus validating the effectiveness and robustness of the approach which combines local planar motion for localization.

\begin{table*}[tbp]
	\caption{Comparison of success rates on OpenLORIS-scene dataset.}
	\begin{center}
			\resizebox{1.0\textwidth}{!}{
			\begin{threeparttable}
				\begin{tabular}{llcccccc}
					\Xhline{1pt}
					\multirow{3}{*}{} & session      & home2                     & home4                     & home5                   & corridor2               & corridor3             & corridor4                 \\
					&m                            & 0.25 / 0.5 / 1.0      & 0.25 / 0.5 / 1.0          & 0.25 / 0.5 / 1.0         & 0.25 / 0.5 / 1.0         & 0.25 / 0.5 / 1.0        & 0.25 / 0.5 / 1.0      \\
					&degree                          & 10 / 10 / 20             & 10 / 10 / 20             & 10 / 10 / 20             & 10 / 10 / 20             & 10 / 10  /20          & 10 / 10 / 20             \\
					\Xhline{1pt}
					&8p8p             & 18.49 / 28.74 / 44.39                        & 9.33 / 18.53 / 39.24 & 56.61 / 66.32 / 77.33 & 2.57 / 5.36 / 11.62 & 0.48 / 1.62 / 4.77  &   3.34 / 5.61 / 10.80   \\
					&5p5p-nister      & 18.53 / 28.68 / 44.96                        & 9.04 /  18.61 / 38.53     & 56.74 / 65.54 / 77.20   & 2.54 / 5.34 / 11.57  & 0.38 / 1.62 / 5.20   & 2.96 / 5.88 / 10.80   \\
					&5p5p-stewenius   & 18.06 / 28.48 / 45.16                        & 9.19 / 18.51 / 38.20     & 55.96 / 64.38 / 76.04   & 2.19 / 5.31 / 11.34 & 0.29 / 1.38 / 5.00   & 3.65 / 6.03 / 11.22   \\

					\multirow{5}{*}{}

					&Tilt-2p2p (ours)   & \textcolor{blue}{22.18} / \textcolor{red}{32.82} / \textcolor{red}{48.03}                        & \textcolor{blue}{11.70} / \textcolor{blue}{22.50} / \textcolor{blue}{43.87}     & \textcolor{blue}{59.82} / \textcolor{red}{69.06} / \textcolor{red}{79.72}   & \textcolor{blue}{2.82} / \textcolor{blue}{6.10} / \textcolor{blue}{12.88} & \textcolor{red}{0.95} / \textcolor{blue}{3.19} / \textcolor{blue}{6.48}   & \textcolor{blue}{4.29} / \textcolor{red}{7.25} / \textcolor{red}{15.14}   \\
					&Tilt-2p1p (ours)    & \textcolor{red}{22.68} / \textcolor{blue}{31.52} / \textcolor{blue}{47.10}                        & \textcolor{red}{12.96} /  \textcolor{red}{23.16} / \textcolor{red}{44.34}     & \textcolor{red}{60.08} / \textcolor{blue}{68.55} / \textcolor{blue}{79.33}   & \textcolor{red}{3.62} / \textcolor{red}{7.79} / \textcolor{red}{13.60}  & \textcolor{blue}{0.91} / \textcolor{red}{3.43} / \textcolor{red}{6.86}   & \textcolor{red}{4.34} / \textcolor{blue}{6.56} / \textcolor{blue}{13.34}   \\ \hline
					\Xhline{1pt}
					& & & & & & & \\
					\Xhline{1pt}
					\multirow{3}{*}{} & session      & corridor5                     & office3                     & office4                   & office5               & office7             & cafe1                 \\
					&m                            & 0.25 / 0.5 / 1.0      & 0.25 / 0.5 / 1.0      & 0.25 / 0.5 / 1.0      & 0.25 / 0.5 / 1.0      & 0.25 / 0.5 / 1.0      & 0.25 / 0.5 / 1.0      \\
					&degree                          & 10 / 10 / 20             & 10 / 10 / 20             & 10 / 10 / 20             & 10 / 10 / 20             & 10 / 10  /20             & 10 / 10 / 20             \\
					\Xhline{1pt}
					&8p8p & 9.04 / 16.66 / 31.02 & 31.74 / 31.74 / 40.17 & 51.90 / 59.03 / 66.05 & 62.24 / 76.09 /  89.43 & 50.22 / 67.67 / 81.76 & 6.91 / 12.29 / 22.09   \\
					&5p5p-nister      & 9.16 / 16.48 / 30.33 & 31.46 / 31.74 / 40.73 & 52.36 / 59.15 / 66.97 & 62.24 / 76.46 / 89.24 & 50.31 / 67.14 / 81.67 & 7.09 / 12.23 / 21.85  \\
					&5p5p-stewenius  & 9.04 / 16.36 / 30.61 & 31.74 / 32.30 / 41.01 & 52.36 / 59.61 / 67.09 & 62.37 /  76.46 / 88.86 & 49.69 / 67.49 / 81.94 & 7.09 / 12.17 / 21.91   \\

					\multirow{5}{*}{}

					&Tilt-2p2p (ours)   & \textcolor{blue}{11.85} / \textcolor{red}{21.29} / \textcolor{red}{34.47}                        & \textcolor{red}{32.50} / \textcolor{blue}{33.89} / \textcolor{red}{43.33}     & \textcolor{red}{56.21} / \textcolor{red}{60.80} / \textcolor{blue}{68.74}   & \textcolor{blue}{66.46} / \textcolor{blue}{79.23} / \textcolor{red}{90.56} & \textcolor{red}{54.51} / \textcolor{blue}{69.76} / \textcolor{red}{83.17}   & \textcolor{blue}{7.49} / \textcolor{blue}{13.93} / \textcolor{blue}{23.95}   \\
					&Tilt-2p1p (ours)     & \textcolor{red}{12.27} / \textcolor{blue}{20.90} / \textcolor{blue}{33.90}  & \textcolor{blue}{32.22} /  \textcolor{red}{34.44} / \textcolor{blue}{43.06}     & \textcolor{blue}{52.41} / \textcolor{blue}{59.89} / \textcolor{red}{69.31}   & \textcolor{red}{69.04} / \textcolor{red}{79.55} / \textcolor{blue}{89.93}  & \textcolor{blue}{54.34} / \textcolor{red}{70.73} / \textcolor{blue}{82.91}   & \textcolor{red}{8.08} / \textcolor{red}{14.40} / \textcolor{red}{24.18}   \\
					 \hline
					\Xhline{1pt}
					
				\end{tabular}
				\begin{tablenotes}
					\footnotesize
					\item[1] Considering the map coverage of the environment, the map sequences for each session are as follows: home1 and home3, corridor1, office1, office2 and office6, cafe2.
					
				\end{tablenotes}
		\end{threeparttable}}
	\end{center}
	\label{table.openloristable}
\end{table*}

\begin{figure*}[tbp]
    \centering
    \includegraphics[width=1.0\textwidth]{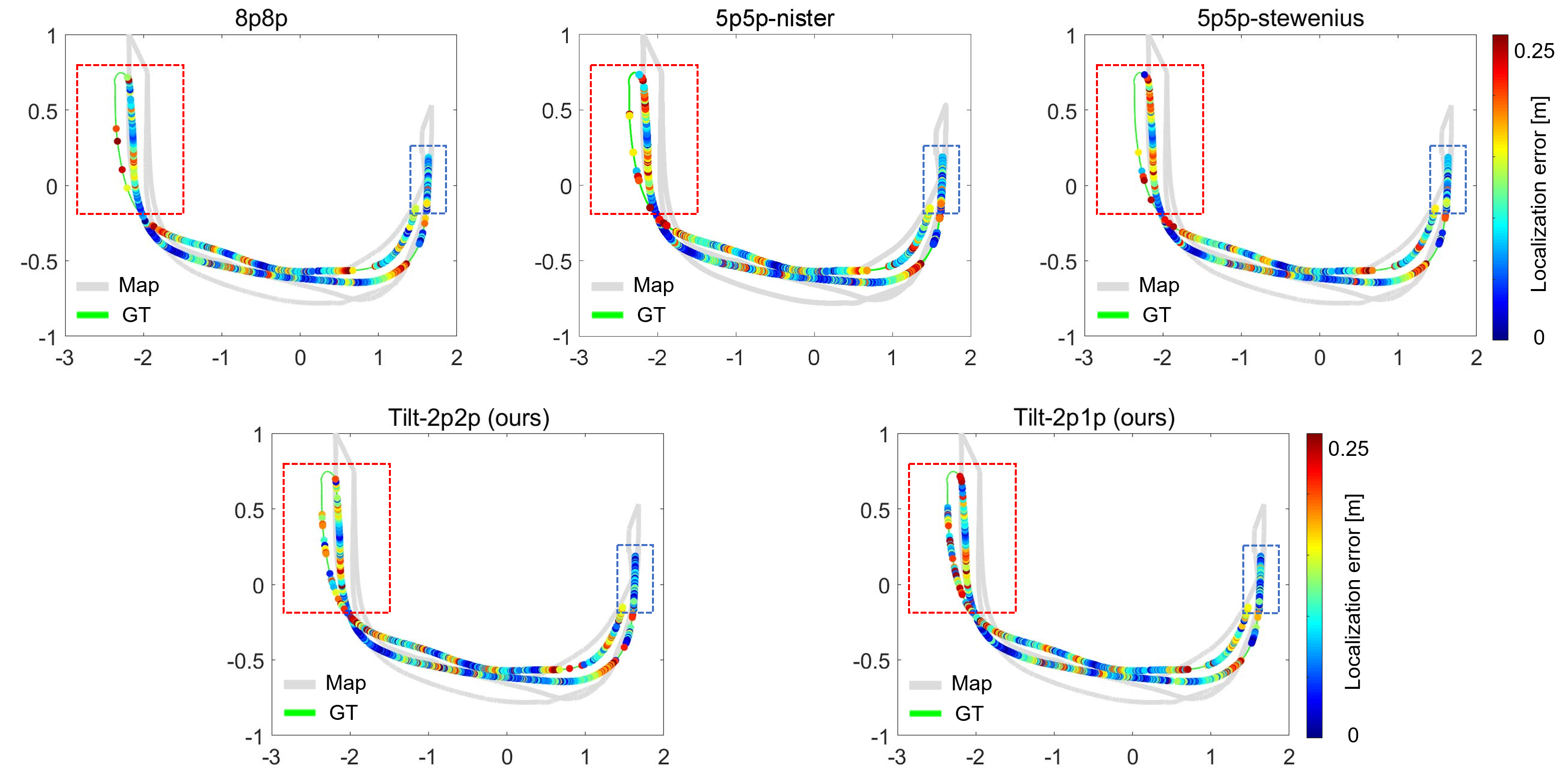}
    \caption{Comparison of estimation errors on whole trajectory of \textit{office5}. The color bar indicates different localization errors. The grey line represents the database trajectory and the green line indicates the ground truth of the query session. The lower the overlap between the gray and green lines, the lower the map coverage, indicating a higher difficulty in localization. As highlighted by the red dashed lines, the proposed methods demonstrate greater robustness in challenging localization scenarios. Additionally, the blue dashed box indicates that the proposed methods achieve higher localization accuracy in areas where map overlap is high. }
    \label{fig.traj}
\end{figure*}

\textbf{Error on Whole Trajectory}: In order to clearly compare the localization accuracy of different methods, we utilize a color bar to display the localization errors of the algorithms on the ground truth trajectory, as illustrated in Fig. \ref{fig.traj}. It is important to note that the localization is performed on each query image without sequential odometry assistance, and only cases with absolute translation error lower than 0.25m are presented on the trajectory. We select the \textit{office5} sequence with the lowest localization difficulty to demonstrate the accuracy comparison. The bluer the color, the smaller the localization error, and higher quantity represents more successful localizations. The results demonstrate that the proposed algorithms achieve superior accuracy and robustness compared to the comparative methods across the entire trajectory, which is consistent with validation using synthetic data presented in Sec. \ref{exp.sim}.

\begin{figure*}[tbp]
    \centering
    \includegraphics[width=1.0\textwidth]{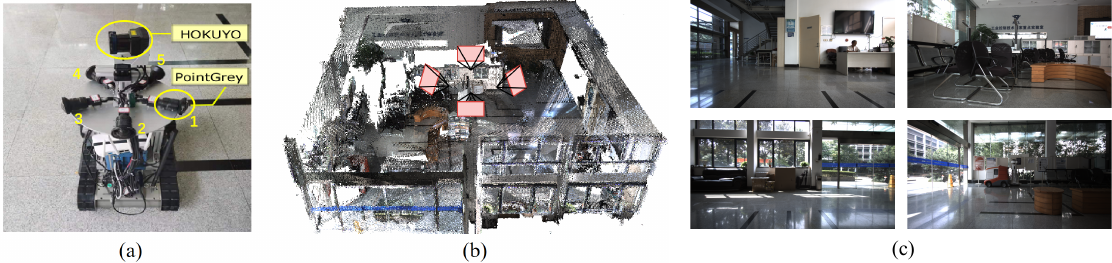}
    \caption{(a) The platform used for Lobby dataset collection. (b) Illustration of the Lobby environment and camera positions to get the database images in (c). }
    \label{fig.hall}
\end{figure*}

\begin{figure}[tbp]
  \centering

  \subfigure[Case 1 with outlier rate of 46\%.]{
            \includegraphics[width=0.46\textwidth]{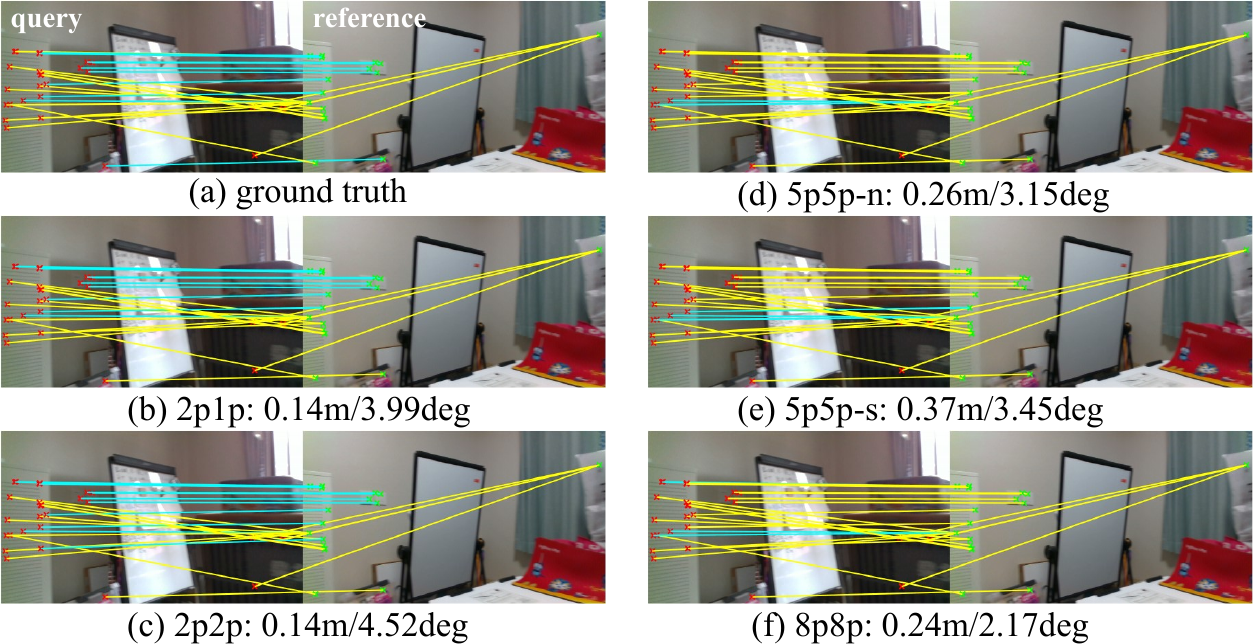}
        }
  \subfigure[Case 2 with only 6 inliers.]{
            \includegraphics[width=0.46\textwidth]{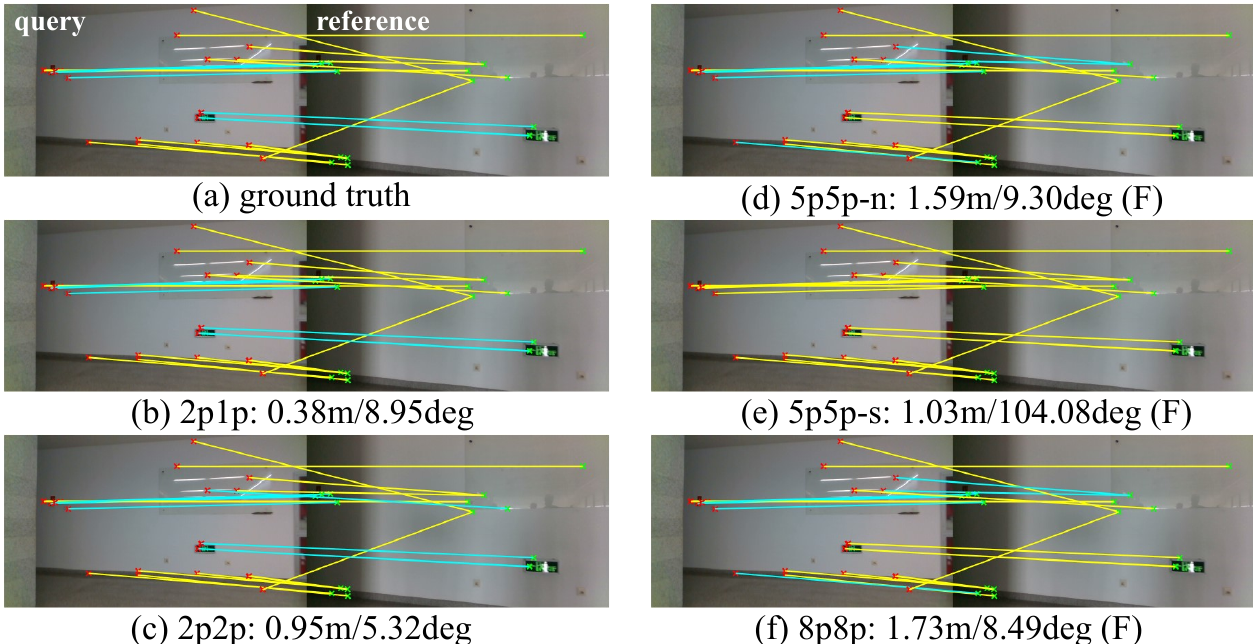}
        }
  \caption{Comparison of identified inliers by different algorithms. The correspondences between the query and reference images are indicated by lines. The identified \textcolor{cyan}{inliers} that support the pose estimation by the corresponding algorithm are highlighted in {cyan}. The estimation errors of different methods are also indicated, with F marking a failure to localize.}
  \label{fig.cases}
\end{figure}

\textbf{Case Study}: We take cases with relatively extreme outlier rates as examples to provide an intuitive demonstration of the advantage of the proposed motion property-aided algorithms against general 6DoF methods, as shown in Fig. \ref{fig.cases}. The lines crossing the query image on the left and the reference image on the right represent matched feature correspondences. Among them, the cyan lines indicate inliers that support the corresponding poses of the ground truth (top left) and the estimated poses of each method.

The first case displays a localization example with an outlier rate of 46\%, where the proposed methods identify almost all inliers and achieve the highest localization accuracy. The second case presents a more challenging scenario where the scene is textureless, resulting in poor feature matching performance with only 6 inliers. As a result, only the proposed 2p1p method, which requires a minimal number of features for pose estimation, achieves high-precision localization. The 2p2p method successfully localizes the scene but with larger errors, leading to the inclusion of an outlier among the identified inliers. In contrast, all comparative methods fail to achieve successful localization as they consider a large number of spurious matches that are incorrectly identified as inliers.

\subsection{Experiments on Lobby}
To further verify the effectiveness of the proposed methods in challenging localization scenarios, we perform validation on a self-collected Lobby dataset \cite{jiao2022leveraging}. The dataset is collected by a mobile robot equipped with five pointgrey cameras and a 2D LiDAR as shown in Fig. \ref{fig.hall} (a). More details about the hardwares can be found in \cite{jiao2022leveraging}. The Lobby is a 11m $\times$ 11m square with two walls made up of large French windows, so the environment is subject to significant changes in external lighting during the day and artificial lighting at night. The static obstacles in the environment mainly include some tables, chairs and posters, while the dynamic obstacles are the moving pedestrians. The data in the Lobby dataset contains environmental changes from afternoon to evening.

\begin{figure}[tbp]
    \centering
    \includegraphics[width=0.45\textwidth]{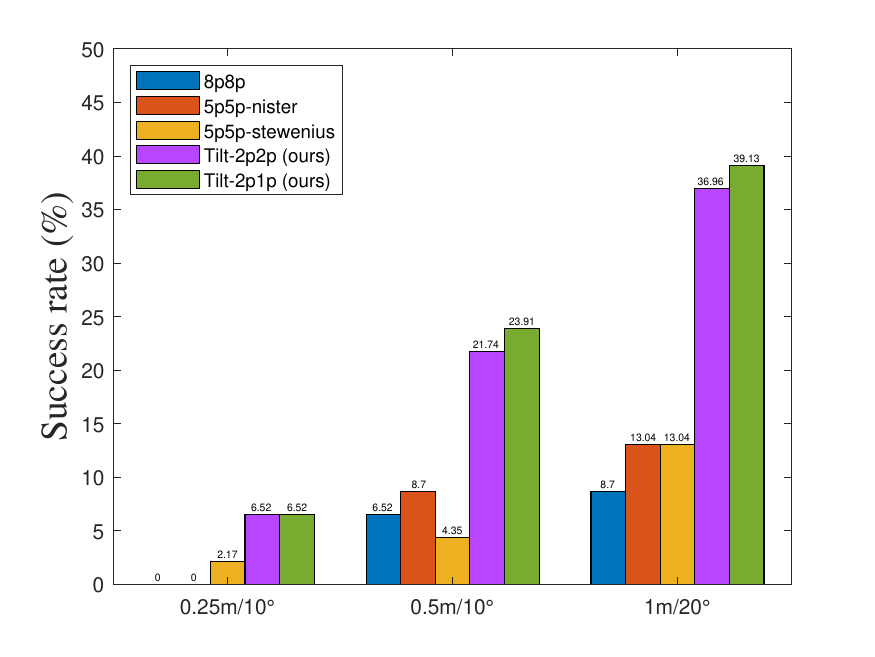}
    \caption{{Comparison of success rates on Lobby dataset.}}
    \label{fig.hallsuc}
\end{figure}

There are total 50 images with a resolution of 1024$\times$540 in the dataset and we select four images facing the four walls to form the database as shown in Fig. \ref{fig.hall} (b) and (c), and the remaining 46 images as query images to perform evaluation. The ground truth of each image is obtained by aligning the dense synchronized 3D scans obtained by rotating the 2D LiDAR for 360 degree.

The localization performance is shown in Fig \ref{fig.hallsuc}. The experimental results indicate that the proposed methods achieve significantly higher localization success rate than the comparative methods under all error thresholds. This is mainly due to the high difficulty of localization in the Lobby scene, with fewer feature matches and a higher proportion of outliers. The result is consistent with that obtained in the experiment conducted in the challenging scenes of OpenLORIS, which once again verifies the effectiveness of the proposed method in robust localization under long-term changing environments.

\section{Conclusion}
In this paper, we propose a multiple checking-based 3D model-free robust visual localization framework and provide thorough validations through simulations and real-world experiments to demonstrate its effectiveness. The key insight of our approach is to utilize the local planar motion property both in feature matching and pose estimation, which is well-suited for general ground-moving robots. In addition, incorporating the motion property into the triangulation stage further reduces the minimal number of required features in pose estimation and enhances the robustness of the proposed system. The experiments reveal a significant improvement in the localization success rate of the proposed system, particularly in challenging localization scenarios. In the future, we plan to incorporate the minimal solution into the back-end of an end-to-end learning-based visual localization pipeline and capitalize on the robustness of pose estimation to improve feature extraction in the front-end.

%
%


\bibliographystyle{IEEEtran}
\bibliography{IEEEabrv,bare_jrnl_new_sample4} 

\begin{thebibliography}{10}
\providecommand{\url}[1]{#1}
\csname url@rmstyle\endcsname
\providecommand{\newblock}{\relax}
\providecommand{\bibinfo}[2]{#2}
\providecommand\BIBentrySTDinterwordspacing{\spaceskip=0pt\relax}
\providecommand\BIBentryALTinterwordstretchfactor{4}
\providecommand\BIBentryALTinterwordspacing{\spaceskip=\fontdimen2\font plus
\BIBentryALTinterwordstretchfactor\fontdimen3\font minus
  \fontdimen4\font\relax}
\providecommand\BIBforeignlanguage[2]{{%
\expandafter\ifx\csname l@#1\endcsname\relax
\typeout{** WARNING: IEEEtran.bst: No hyphenation pattern has been}%
\typeout{** loaded for the language `#1'. Using the pattern for}%
\typeout{** the default language instead.}%
\else
\language=\csname l@#1\endcsname
\fi
#2}}

\bibitem{cheng2019robust}
J.~Cheng, C.~Wang, and M.~Q.-H. Meng, ``Robust visual localization in dynamic
  environments based on sparse motion removal,'' \emph{IEEE Transactions on
  Automation Science and Engineering}, vol.~17, no.~2, pp. 658--669, 2019.

\bibitem{cheng2020improving}
J.~Cheng, H.~Zhang, and M.~Q.-H. Meng, ``Improving visual localization accuracy
  in dynamic environments based on dynamic region removal,'' \emph{IEEE
  Transactions on Automation Science and Engineering}, vol.~17, no.~3, pp.
  1585--1596, 2020.

\bibitem{li2017indoor}
R.~Li, Q.~Liu, J.~Gui, D.~Gu, and H.~Hu, ``Indoor relocalization in challenging
  environments with dual-stream convolutional neural networks,'' \emph{IEEE
  Transactions on Automation Science and Engineering}, vol.~15, no.~2, pp.
  651--662, 2017.

\bibitem{santoso2016visual}
F.~Santoso, M.~A. Garratt, and S.~G. Anavatti, ``Visual--inertial navigation
  systems for aerial robotics: Sensor fusion and technology,'' \emph{IEEE
  Transactions on Automation Science and Engineering}, vol.~14, no.~1, pp.
  260--275, 2016.

\bibitem{jiao2021deterministic}
Y.~Jiao, Y.~Wang, X.~Ding, M.~Wang, and R.~Xiong, ``Deterministic optimality
  for robust vehicle localization using visual measurements,'' \emph{IEEE
  Transactions on Intelligent Transportation Systems}, 2021.

\bibitem{zhang2023toward}
Z.~Zhang, Y.~Song, S.~Huang, R.~Xiong, and Y.~Wang, ``Toward consistent and
  efficient map-based visual-inertial localization: Theory framework and filter
  design,'' \emph{IEEE Transactions on Robotics}, 2023.

\bibitem{schonberger2016structure}
J.~L. Schonberger and J.-M. Frahm, ``Structure-from-motion revisited,'' in
  \emph{Proceedings of the IEEE conference on computer vision and pattern
  recognition}, 2016, pp. 4104--4113.

\bibitem{kendall2015posenet}
A.~Kendall, M.~Grimes, and R.~Cipolla, ``Posenet: A convolutional network for
  real-time 6-dof camera relocalization,'' in \emph{Proceedings of the IEEE
  international conference on computer vision}, 2015, pp. 2938--2946.

\bibitem{brachmann2017dsac}
E.~Brachmann, A.~Krull, S.~Nowozin, J.~Shotton, F.~Michel, S.~Gumhold, and
  C.~Rother, ``Dsac-differentiable ransac for camera localization,'' in
  \emph{Proceedings of the IEEE conference on computer vision and pattern
  recognition}, 2017, pp. 6684--6692.

\bibitem{balntas2018relocnet}
V.~Balntas, S.~Li, and V.~Prisacariu, ``Relocnet: Continuous metric learning
  relocalisation using neural nets,'' in \emph{Proceedings of the European
  Conference on Computer Vision (ECCV)}, 2018, pp. 751--767.

\bibitem{lowe2004distinctive}
D.~G. Lowe, ``Distinctive image features from scale-invariant keypoints,''
  \emph{International journal of computer vision}, vol.~60, no.~2, pp. 91--110,
  2004.

\bibitem{detone2018superpoint}
D.~DeTone, T.~Malisiewicz, and A.~Rabinovich, ``Superpoint: Self-supervised
  interest point detection and description,'' in \emph{Proceedings of the IEEE
  conference on computer vision and pattern recognition workshops}, 2018, pp.
  224--236.

\bibitem{rublee2011orb}
E.~Rublee, V.~Rabaud, K.~Konolige, and G.~Bradski, ``Orb: An efficient
  alternative to sift or surf,'' in \emph{2011 International conference on
  computer vision}.\hskip 1em plus 0.5em minus 0.4em\relax Ieee, 2011, pp.
  2564--2571.

\bibitem{jiao20202}
Y.~Jiao, Y.~Wang, X.~Ding, B.~Fu, S.~Huang, and R.~Xiong, ``2-entity ransac for
  robust visual localization: Framework, methods and verifications,''
  \emph{IEEE Transactions on Industrial Electronics}, 2020.

\bibitem{gao2003complete}
X.-S. Gao, X.-R. Hou, J.~Tang, and H.-F. Cheng, ``Complete solution
  classification for the perspective-three-point problem,'' \emph{IEEE
  transactions on pattern analysis and machine intelligence}, vol.~25, no.~8,
  pp. 930--943, 2003.

\bibitem{lepetit2009epnp}
V.~Lepetit, F.~Moreno-Noguer, and P.~Fua, ``Epnp: An accurate o (n) solution to
  the pnp problem,'' \emph{International journal of computer vision}, vol.~81,
  no.~2, p. 155, 2009.

\bibitem{arandjelovic2016netvlad}
R.~Arandjelovic, P.~Gronat, A.~Torii, T.~Pajdla, and J.~Sivic, ``Netvlad: Cnn
  architecture for weakly supervised place recognition,'' in \emph{Proceedings
  of the IEEE conference on computer vision and pattern recognition}, 2016, pp.
  5297--5307.

\bibitem{lowry2015visual}
S.~Lowry, N.~S{\"u}nderhauf, P.~Newman, J.~J. Leonard, D.~Cox, P.~Corke, and
  M.~J. Milford, ``Visual place recognition: A survey,'' \emph{ieee
  transactions on robotics}, vol.~32, no.~1, pp. 1--19, 2015.

\bibitem{masone2021survey}
C.~Masone and B.~Caputo, ``A survey on deep visual place recognition,''
  \emph{IEEE Access}, vol.~9, pp. 19\,516--19\,547, 2021.

\bibitem{hartley2003multiple}
R.~Hartley and A.~Zisserman, \emph{Multiple view geometry in computer
  vision}.\hskip 1em plus 0.5em minus 0.4em\relax Cambridge university press,
  2003.

\bibitem{nister2004efficient}
D.~Nist{\'e}r, ``An efficient solution to the five-point relative pose
  problem,'' \emph{IEEE transactions on pattern analysis and machine
  intelligence}, vol.~26, no.~6, pp. 756--770, 2004.

\bibitem{stewenius2006recent}
H.~Stewenius, C.~Engels, and D.~Nist{\'e}r, ``Recent developments on direct
  relative orientation,'' \emph{ISPRS Journal of Photogrammetry and Remote
  Sensing}, vol.~60, no.~4, pp. 284--294, 2006.

\bibitem{fischler2014readings}
M.~A. Fischler and O.~Firschein, \emph{Readings in computer vision: issues,
  problem, principles, and paradigms}.\hskip 1em plus 0.5em minus 0.4em\relax
  Elsevier, 2014.

\bibitem{cao2014minimal}
S.~Cao and N.~Snavely, ``Minimal scene descriptions from structure from motion
  models,'' in \emph{Proceedings of the IEEE Conference on Computer Vision and
  Pattern Recognition}, 2014, pp. 461--468.

\bibitem{choudhary2012visibility}
S.~Choudhary and P.~Narayanan, ``Visibility probability structure from sfm
  datasets and applications,'' in \emph{Computer Vision--ECCV 2012: 12th
  European Conference on Computer Vision, Florence, Italy, October 7-13, 2012,
  Proceedings, Part V 12}.\hskip 1em plus 0.5em minus 0.4em\relax Springer,
  2012, pp. 130--143.

\bibitem{geppert2019efficient}
M.~Geppert, P.~Liu, Z.~Cui, M.~Pollefeys, and T.~Sattler, ``Efficient 2d-3d
  matching for multi-camera visual localization,'' in \emph{2019 International
  Conference on Robotics and Automation (ICRA)}.\hskip 1em plus 0.5em minus
  0.4em\relax IEEE, 2019, pp. 5972--5978.

\bibitem{li2016worldwide}
Y.~Li, N.~Snavely, D.~P. Huttenlocher, and P.~Fua, ``Worldwide pose estimation
  using 3d point clouds,'' \emph{Large-Scale Visual Geo-Localization}, pp.
  147--163, 2016.

\bibitem{brachmann2018learning}
E.~Brachmann and C.~Rother, ``Learning less is more-6d camera localization via
  3d surface regression,'' in \emph{Proceedings of the IEEE conference on
  computer vision and pattern recognition}, 2018, pp. 4654--4662.

\bibitem{donoser2014discriminative}
M.~Donoser and D.~Schmalstieg, ``Discriminative feature-to-point matching in
  image-based localization,'' in \emph{Proceedings of the IEEE Conference on
  Computer Vision and Pattern Recognition}, 2014, pp. 516--523.

\bibitem{kendall2017geometric}
A.~Kendall and R.~Cipolla, ``Geometric loss functions for camera pose
  regression with deep learning,'' in \emph{Proceedings of the IEEE conference
  on computer vision and pattern recognition}, 2017, pp. 5974--5983.

\bibitem{sarlin2019coarse}
P.-E. Sarlin, C.~Cadena, R.~Siegwart, and M.~Dymczyk, ``From coarse to fine:
  Robust hierarchical localization at large scale,'' in \emph{Proceedings of
  the IEEE/CVF Conference on Computer Vision and Pattern Recognition}, 2019,
  pp. 12\,716--12\,725.

\bibitem{irschara2009structure}
A.~Irschara, C.~Zach, J.-M. Frahm, and H.~Bischof, ``From structure-from-motion
  point clouds to fast location recognition,'' in \emph{2009 IEEE Conference on
  Computer Vision and Pattern Recognition}.\hskip 1em plus 0.5em minus
  0.4em\relax IEEE, 2009, pp. 2599--2606.

\bibitem{sarlin2018leveraging}
P.-E. Sarlin, F.~Debraine, M.~Dymczyk, R.~Siegwart, and C.~Cadena, ``Leveraging
  deep visual descriptors for hierarchical efficient localization,'' in
  \emph{Conference on Robot Learning}, 2018, pp. 456--465.

\bibitem{ortin2001indoor}
D.~Ortin and J.~M.~M. Montiel, ``Indoor robot motion based on monocular
  images,'' \emph{Robotica}, vol.~19, no.~3, pp. 331--342, 2001.

\bibitem{chou20152}
C.~C. Chou and C.-C. Wang, ``2-point ransac for scene image matching under
  large viewpoint changes,'' in \emph{2015 IEEE International Conference on
  Robotics and Automation (ICRA)}.\hskip 1em plus 0.5em minus 0.4em\relax IEEE,
  2015, pp. 3646--3651.

\bibitem{hong2016improved}
S.~Hong, J.~S. Lee, and T.-Y. Kuc, ``Improved algorithm to estimate the
  rotation angle between two images by using the two-point correspondence
  pairs,'' \emph{Electronics Letters}, vol.~52, no.~5, pp. 355--357, 2016.

\bibitem{scaramuzza2009real}
D.~Scaramuzza, F.~Fraundorfer, and R.~Siegwart, ``Real-time monocular visual
  odometry for on-road vehicles with 1-point ransac,'' in \emph{2009 IEEE
  International conference on robotics and automation}.\hskip 1em plus 0.5em
  minus 0.4em\relax Ieee, 2009, pp. 4293--4299.

\bibitem{zhou2020learn}
Q.~Zhou, T.~Sattler, M.~Pollefeys, and L.~Leal-Taixe, ``To learn or not to
  learn: Visual localization from essential matrices,'' in \emph{2020 IEEE
  International Conference on Robotics and Automation (ICRA)}.\hskip 1em plus
  0.5em minus 0.4em\relax IEEE, 2020, pp. 3319--3326.

\bibitem{jiao2022leveraging}
Y.~Jiao, Q.~Zhang, Q.~Chen, B.~Fu, F.~Han, Y.~Wang, and R.~Xiong, ``Leveraging
  local planar motion property for robust visual matching and localization,''
  \emph{IEEE Robotics and Automation Letters}, vol.~7, no.~3, pp. 7589--7596,
  2022.

\bibitem{kneip2014opengv}
L.~Kneip and P.~Furgale, ``Opengv: A unified and generalized approach to
  real-time calibrated geometric vision,'' in \emph{2014 IEEE International
  Conference on Robotics and Automation (ICRA)}.\hskip 1em plus 0.5em minus
  0.4em\relax IEEE, 2014, pp. 1--8.

\bibitem{jiao2020robust}
Y.~Jiao, L.~Liu, B.~Fu, X.~Ding, M.~Wang, Y.~Wang, and R.~Xiong, ``Robust
  localization for planar moving robot in changing environment: A perspective
  on density of correspondence and depth,'' in \emph{2021 IEEE International
  Conference on Robotics and Automation (ICRA)}.\hskip 1em plus 0.5em minus
  0.4em\relax IEEE, 2021, pp. 4006--4012.

\bibitem{sattler2018benchmarking}
T.~Sattler, W.~Maddern, C.~Toft, A.~Torii, L.~Hammarstrand, E.~Stenborg,
  D.~Safari, M.~Okutomi, M.~Pollefeys, J.~Sivic, \emph{et~al.}, ``Benchmarking
  6dof outdoor visual localization in changing conditions,'' in
  \emph{Proceedings of the IEEE Conference on Computer Vision and Pattern
  Recognition}, 2018, pp. 8601--8610.

\bibitem{shi2020we}
X.~Shi, D.~Li, P.~Zhao, Q.~Tian, Y.~Tian, Q.~Long, C.~Zhu, J.~Song, F.~Qiao,
  L.~Song, \emph{et~al.}, ``Are we ready for service robots? the
  openloris-scene datasets for lifelong slam,'' in \emph{2020 IEEE
  International Conference on Robotics and Automation (ICRA)}.\hskip 1em plus
  0.5em minus 0.4em\relax IEEE, 2020, pp. 3139--3145.

\bibitem{revaud2019r2d2}
J.~Revaud, P.~Weinzaepfel, C.~De~Souza, N.~Pion, G.~Csurka, Y.~Cabon, and
  M.~Humenberger, ``R2d2: repeatable and reliable detector and descriptor,'' in
  \emph{Conference on Neural Information Processing Systems (NeurIPS)}, 2019.

\end{thebibliography}

\end{document}